\pdfoutput=1

\documentclass[11pt]{article}

\usepackage[preprint]{acl}

\usepackage{times}
\usepackage{latexsym}

\usepackage[T1]{fontenc}

\usepackage[utf8]{inputenc}

\usepackage{microtype}

\usepackage{inconsolata}

\usepackage{graphicx}
\usepackage{booktabs}
\usepackage{multirow}
\usepackage{colortbl}
\usepackage{xcolor}
\usepackage{makecell}

\usepackage{graphicx}
\usepackage{tcolorbox}
\usepackage{xcolor}
\usepackage{enumitem}
\usepackage{subfigure}   
\usepackage{float}  
\usepackage{bm}
\usepackage{multirow}
\usepackage{xspace}
\usepackage{enumitem}
\usepackage{appendix}
\usepackage{hyperref}
\usepackage{xcolor}
\usepackage{graphicx}
\usepackage{subcaption}
\usepackage{booktabs} 
\usepackage{multirow}
\usepackage{enumitem}
\usepackage{balance}
\usepackage{makecell}
\usepackage{threeparttable}
\usepackage{amsmath,amsfonts,mathtools}
\usepackage{mathrsfs}
\usepackage{float}
\usepackage{graphicx}
\usepackage{algorithm,algorithmic}
\usepackage{makecell}
\usepackage{booktabs}
\usepackage{pifont}

\usepackage{multirow}
\usepackage{enumitem}
\usepackage{balance}
\usepackage{threeparttable}
\usepackage{colortbl}
\usepackage{mathrsfs}
\usepackage{float}
\usepackage{graphicx}
\usepackage{subfigure}
\usepackage{hyperref}
\usepackage{tabularx, booktabs}
\usepackage{tikz}
\usepackage{arydshln}
\usepackage{CJKutf8}

\newcommand{\mname}{Parenting\xspace}
\title{\mname: Optimizing Knowledge Selection of Retrieval-Augmented Language Models with Parameter Decoupling and Tailored Tuning}

\author{Yongxin Xu$^{1,2}$\thanks{Equal contribution.}, Ruizhe Zhang$^{1,2}$\footnotemark[1], Xinke Jiang$^{1,2}$\footnotemark[1], Yujie Feng$^{7}$, Yuzhen Xiao$^{1,2}$, \\
\textbf{Xinyu Ma}$^{1,2}$\textbf{,} \textbf{Runchuan Zhu}$^{1,2}$\textbf{,} \textbf{Xu Chu}$^{1,2,4,5}$\textbf{,} \textbf{Junfeng Zhao}$^{1,2,6}$\thanks{Corresponding author.}, \textbf{Yasha Wang}$^{2,3,4}$\footnotemark[2]\\
    $^{1}$ School of Computer Science and School of Software \& Microelectronics, Peking University\\
    $^{2}$ Key Laboratory of High Confidence Software Technologies, Ministry of Education\\ 
    $^{3}$ National Engineering Research Center For Software Engineering, Peking University\\
    $^{4}$ Peking University Information Technology Institute (Tianjin Binhai)\\
    $^{5}$ Center on Frontiers of Computing Studies, Peking University\\
    $^{6}$ Big Data Technology Research Center, Nanhu Laboratory\\
    $^{7}$ Department of Computing, The Hong Kong Polytechnic University\\
    \small
    \{xuyx, nostradamus, xinkejiang\}@stu.pku.edu.cn, \{zhaojf, wangyasha\}@pku.edu.cn
}

\begin{document}
\maketitle
\begin{abstract}
Retrieval-Augmented Generation (RAG) offers an effective solution to the issues faced by Large Language Models (LLMs) in hallucination generation and knowledge obsolescence by incorporating externally retrieved knowledge.
However, existing methods lack effective control mechanisms for integrating internal and external knowledge.
Inspired by human cognitive processes, we propose \mname, a novel framework that decouples, identifies, and purposefully optimizes parameter subspaces related to adherence and robustness.
Specifically, \mname utilizes a key parameter mining method that combines forward and backward propagation signals to localize subspaces representing different capabilities. 
Then, \mname employs a type-tailored tuning strategy, applying specific and appropriate optimizations to different subspaces, aiming to achieve a balanced enhancement of both adherence and robustness.
Extensive experiments on various datasets and models validate the effectiveness and generalizability of our method.
Our code is available at \url{https://github.com/Nostradamus4869/Parenting}.
\end{abstract}

\section{Introduction}
Large Language Models (LLMs) have demonstrated exceptional capabilities, achieving state-of-the-art performance on various tasks~\cite{brown2020language,chowdhery2023palm,bubeck2023sparks,xu2025dearllm,ma2025dressing,feng2023towards,feng2025recurrent}. 
Despite their success, they still face notable challenges, particularly in generating hallucinations and dealing with outdated knowledge~\cite{gao2023retrieval}. 
Retrieval-Augmented Generation (RAG) has emerged as a promising approach to mitigate these issues~\cite{peng2023check,ren2023investigating,lewis2020retrieval,jiang2024tc}.
By integrating external information relevant to a specific query, RAG enhances the generative process with supplementary non-parametric knowledge.
However, typical RAG frameworks lack effective control mechanisms for managing internal and external knowledge~\cite{li2023large}, which presents two main challenges: 
First, the conflicts between external knowledge and the internal memory of LLMs can, in certain cases, prevent the model from effectively following external evidence to produce accurate responses~\cite{wu2024faithful,li2023large}.
Secondly, the inherent imperfections of retrieval mechanisms mean that the retrieved contexts might include irrelevant noises~\cite{creswellselection}, which can mislead the LLMs, leading to degraded performance~\cite{fang2024enhancing,xu2024unsupervised}. 

To address the above issues, some approaches optimize knowledge selection in the RAG process through carefully designed prompts~\cite{zhou2023context}. However, such methods do not fundamentally improve LLMs' ability to integrate external knowledge, leading to suboptimal outcomes in certain situations.
Other methods focus on altering the behavioral patterns of LLMs through training techniques such as instruction tuning~\cite{li2023large,fang2024enhancing,xu2024unsupervised}. Nevertheless, they lack differentiation in supervisory signals, leading to significant learning variance.
On the one hand, an excessive emphasis on adhering to context can lead models to pay attention to noisy information. On the other hand, prioritizing resistance to noise might cause models to overlook critical evidence present within the context~\cite{wu2024clasheval}.
In summary, how to establish an effective control mechanism for managing both internal and external knowledge within RAG remains an unresolved problem.

The human brain is comprised of multiple functional regions, each tasked with distinct cognitive and physiological roles, with some regions being integral to handling complex tasks~\cite{hawrylycz2012anatomically}. For example, the \textit{mirror neuron system} enhances learning through observation and imitation~\cite{rizzolatti2004mirror}, while the \textit{hippocampus} plays a crucial role in retrieving knowledge from stored memories to solve problems~\cite{olton1979hippocampus}.
Inspired by this, a natural question arises:

\textbf{\textit{Can we pinpoint specific subspaces in LLMs' parameter space linked to adhering to contextual evidence (adherence) and resisting contextual noise (robustness), akin to how the human brain localizes different cognitive functions?}}

\begin{figure}[t]
    \centering
    \includegraphics[width=1.0\linewidth, height=4cm]{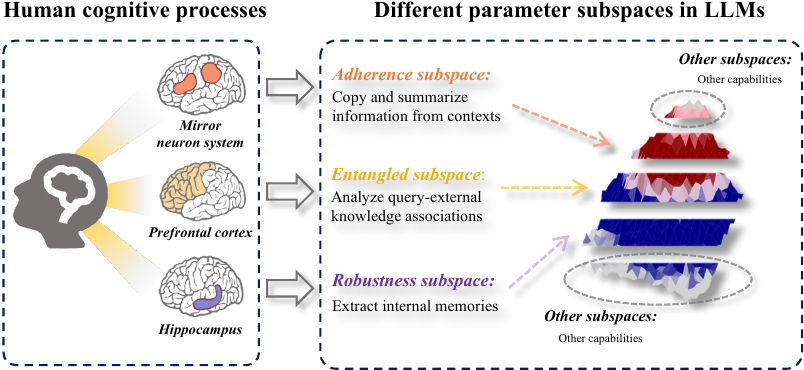}
    \caption{Formal exposition of our critical ideas. Inspired by human cognitive processes, we aim to decouple and localize parameter subspaces linked to distinct abilities, enabling effective behavior control through tailored tuning.}
    \label{fig:introduction}
\end{figure}

To answer this question, 
inspired by ~\citet{feng2024tasl} on identifying key parameters and integrating knowledge from different tasks to alleviate catastrophic forgetting in continual learning, 
we introduce a new perspective, as illustrated in Figure~\ref{fig:introduction}, that aims to decouple and identify parameter subspaces related to \textbf{adherence} and \textbf{robustness}. By designing tailored tuning strategies for these different subspaces, we seek to achieve better control over model behavior.
Although seemingly straightforward, implementing this intuition faces these challenges: 
\textbf{(C1)} How to precisely quantify the correlation between parameters and two capabilities? 
\textbf{(C2)} How to optimize the entangled subspace that is difficult to decouple?

To address these challenges, we propose a framework named \mname, which leverages \textbf{\underline{Par}}ameter D\textbf{\underline{e}}coupli\textbf{\underline{n}}g and \textbf{\underline{T}}ailored Tun\textbf{\underline{ing}} to optimize the internal and external knowledge control mechanisms of Retrieval-Augmented Language Models (RALMs), while boosting both adherence and robustness without compromise. 
Specifically, addressing challenge \textbf{C1}, \mname utilizes a key parameter mining method. 
By constructing a specialized probing dataset, \mname measures each parameter's contribution to specific behaviors by combining the activation levels of pre-trained parameters under various inputs with gradient trajectories observed during the backpropagation process.
Subsequently, through interactive analysis of the parameter importance distribution for different behaviors, we identify two mutually exclusive parameter subspaces: the \textbf{adherence subspace} (analogous to the \textit{mirror neuron system}) and the \textbf{robustness subspace} (similar to the \textit{hippocampus}), along with an \textbf{entangled subspace} that is challenging to decouple. 
Addressing challenge \textbf{C2}, \mname adopts a type-tailored tuning strategy, applying specific optimizations for different subspaces. 
Initially, for the entangled subspace, which consists of parameters critical to both adherence and robustness—similar to the \textit{prefrontal cortex} in humans, which is responsible for perceiving the external environment~\cite{puig2011serotonin}—\mname employs a novel document extraction task to reinforce training, thereby enhancing both adherence and robustness simultaneously.
Concurrently, \mname employs a boundary-controlled fine-tuning strategy to prevent contradictory supervision signals from contaminating the adherence and robustness subspaces. 
To summarize, we highlight our contributions as follows:

\begin{itemize} [leftmargin=*]
    \item We propose a novel insight from the perspective of parameter space to achieve fine-grained decoupling and fine-tuning optimization of adherence and robustness in RALMs.
    \item We propose a key parameter mining method and a type-tailored tuning strategy, which effectively harmonize and enhance the supervisory signals representing adherence and robustness.
    \item We conduct extensive experiments across multiple models and datasets, and the results demonstrate that our approach not only achieves a balance and enhancement of model performance, but also enhances its adaptability and robustness when facing out-of-distribution (OOD) data.
\end{itemize}

\section{Related Work}
\label{section:related work}

\subsection{Retrieval-Augmented Language Models}
RALMs integrate external knowledge bases, allowing the generation process of LLMs to be adjusted based on the most up-to-date and relevant documents or knowledge~\cite{guu2020retrieval}. 
The classic ``Retrieve-Read'' framework leverages the initial input as a query to retrieve relevant information from an external corpus. The retrieved knowledge is then integrated into the input and incorporated into the model's generation process~\cite{gao2023retrieval,fan2024survey}.
For instance, kNN-LM~\cite{khandelwal2019generalization} retrieves external memory using k-nearest neighbors.
To prevent confusion and resource waste caused by excessive retrieval, FLARE~\cite{jiang2023active}, Self-RAG~\cite{asaiself}, and DRAGIN~\cite{su2024dragin} combine LLMs' own information needs to determine when to retrieve and what queries to generate.
Although these approaches are effective, they still face challenges such as confusion caused by conflicts between internal and external knowledge, as well as misleading effects from external noise~\cite{mallen2023not}.

\subsection{Addressing Noisy Contexts and Internal-External Knowledge Conflicts}
After retrieving external knowledge, effectively integrating it with the internal knowledge of LLMs is essential for enhancing the quality of the final output response~\cite{fan2024survey}.
On the one hand, retrieval algorithms cannot achieve complete accuracy, and therefore RALMs will inevitably introduce task-irrelevant noises~\cite{creswellselection}.
To address the above issue, SA-RetRobust~\cite{yoranmaking} enhances the noise robustness of LLMs by introducing an additional fine-tuning step.
Info-RAG~\cite{xu2024unsupervised} and RAAT~\cite{fang2024enhancing} design unsupervised training and adaptive adversarial training strategies, respectively.
On the other hand, models tend to become confused when there is conflict between external knowledge and the internal memory of LLMs~\cite{wu2024faithful}, particularly when the external factual evidence is incoherent or semantically incomplete~\cite{tan2024blinded,xieadaptive}.
To address the aforementioned challenge, prompt-based methods improve LLMs' faithfulness to context through carefully designed prompting strategies~\cite{zhou2023context}.
IRCAN~\cite{shi2024ircan} identifies and reweights adherence-specific neurons.
Decoding-based methods measure differences in output probabilities between conditions with and without context~\cite{shi2024trusting,jin2024tug}.
KAFT~\cite{li2023large} focuses on constructing specific instruction-tuning datasets.
PH3~\cite{jin2024cutting} mitigates knowledge conflicts by pruning negative attention heads using the path patching technique.
KnowPO~\cite{zhang2025knowpo} guides the model to avoid errors in knowledge selection by constructing a knowledge conflict dataset and incorporating preference optimization training.

\section{Task Formulation}
\subsection{Task Definition of RALMs}
Given a natural language query $q$ as input, a standard RALMs typically employs a retriever to retrieve a set of documents $D=\left\{d_1, \ldots, d_k\right\}$ from an external knowledge base $\mathbb{C}$, where $k$ is the window size of the retrieval process.
Then, during the inference process, the retrieved context $D$ is concatenated with the query $q$ and is fed into LLM $\Theta$ to generate the answer $ans_\Theta$, denoted as $ans_\Theta =\Theta\left(q \| D\right)$. Besides, LLMs will use their parametric knowledge $\alpha = \Theta\left(q\right)$ to answer such query without retrieved context. 
We refer to the evidence that can assist in answering $q$ as $d_{golden}$. Next, we define a relevance label $y$ to indicate whether $D$ contains this evidence (i.e., if $d_{golden}\in D$, then $y=1$; otherwise, $y=0$).

Following~\citet{li2023large}, our goal is to enable the fine-tuned LLM $\Theta^\prime$ to derive answers based on retrieved context when it contains valid evidence, even if this evidence conflicts with its internal memory (we refer to such situations as \textbf{conflicting contexts}.). 
Conversely, when the retrieved context contains only semantically similar yet irrelevant noise (consistent with \citet{yoranmaking}, we define them as \textbf{irrelevant contexts}), 
the model should detect the flaws in the context and rely on its internal knowledge to respond. This not only effectively prevents the model from confidently producing incorrect answers due to noisy input, but is also more user-friendly than consistently refusing to answer:
\begin{equation}
\small
\begin{aligned}
ans_{\Theta^\prime} = 
\begin{cases} 
ans_{golden}, & \text{if } y = 1 \\
\alpha, & \text{if } y = 0 
\end{cases},
\end{aligned}
\end{equation}
where $ans_{golden}$ is the correct answer from $d_{golden}$.

\begin{figure*}[t]
  \centering
  \includegraphics[width=1.0\linewidth]{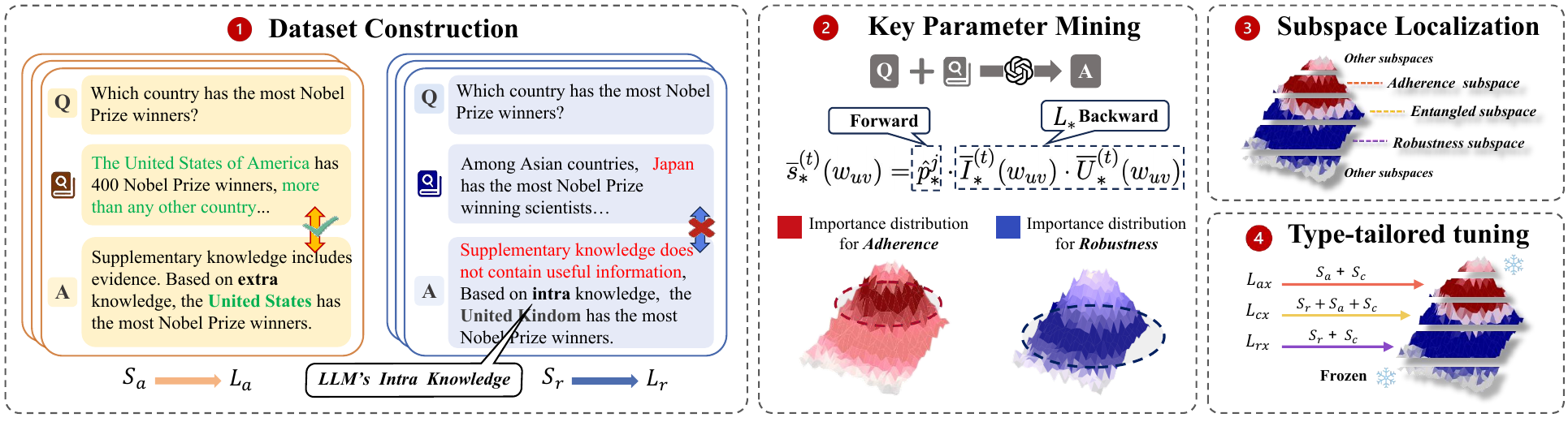}
  \caption{The overview of our proposed \mname.
  }
  \label{fig:method}
\end{figure*}

\subsection{Definition of Parameter Units}
Knowledge organization within LLMs can be described as independent, modular regions, where distinct matrices are typically regarded as fundamental units for knowledge storage~\cite{wang2024knowledge,wang2023knowledge}. 
Therefore, in this paper, we define the fundamental structural elements that store skills, termed ``parameter units'', as individual matrices within the model.
During full parameter fine-tuning, the matrices of the Multi-Head Self-Attention (MHA) and the Feed-Forward Neural Network (FFN) are regarded as basic parameter units. When employing Parameter-Efficient Fine-Tuning (PEFT) techniques~\cite{pmlr-v235-ma24a,lin2024lora} to accelerate training, we consider matrices $A$ and $B$ in LoRA~\cite{hulora} as separate basic parameter units.
A subspace is composed of multiple parameter units.
The model $\Theta$ with $n$ parameter units is denoted as $\mathcal{E}=\left\{e_1, \ldots, e_n\right\}$.
The notations used throughout this paper are provided in Appendix~\ref{sec:app notation}.

\section{Methodology}
\mname includes four components: 
\begin{itemize}[leftmargin=*,noitemsep,topsep=2pt]
\item \textbf{Dataset Construction}, which constructs two supervised fine-tuning (SFT) datasets to stimulate the adherence and robustness of LLMs.
\item \textbf{Key Parameter Mining}, which measures parameter importance by combining forward activations with gradient trajectories.
\item \textbf{Subspace Localization}, which identifies and locates different types of subspaces.
\item \textbf{Type-Tailored Tuning}, which designs specific and appropriate fine-tuning strategies for each type, aiming to achieve a balanced enhancement of adherence and robustness.
\end{itemize}
Figure~\ref{fig:method} presents a detailed overview of our proposed \mname framework, with subsequent subsections detailing each component.

\subsection{Dataset Construction}
\label{section:sft}
To elicit adherence and robustness of LLMs, we design two datasets derived from SQuAD2.0 dataset~\cite{rajpurkar2018knowdontknowunanswerable}, a reading comprehension dataset encompassing multiple general domains, with a substantial corpus of documents and associated question answering (QA) tasks.
We first extract the LLMs' parametric knowledge $\alpha$ using probes for each SQuAD2.0 question. 
Notably, SQuAD2.0 employs manual annotations to indicate whether a document provides an answer to a specific question. 
Based on these annotations, we construct a SFT dataset $S_a$ that promotes adherence and another dataset $S_r$ that enhances robustness.

For dataset $S_a$, we begin by selecting ground truth answers and corresponding evidence documents from SQuAD2.0 that conflict with the parametric knowledge $\alpha$. 
Following~\citet{li2023large}, we further introduce fictional answers and related documents to effectively expand the dataset.
$S_a$ can be represented as $S_a = \{ (q_a^i, D_a^i, ans_a^i) \mid i = 1, 2, \ldots, m_a \}$, where $m_a$ denotes the number of samples in $S_a$. The cross-entropy loss on this dataset is denoted as $\mathcal{L}_a$. 
For dataset $S_r$, regarding the context, we select the manually annotated irrelevant noise documents from the dataset. 
We expect the model to indicate that the context lacks the clues necessary to answer the question and to generate its original parametric knowledge $\alpha$, the cross-entropy loss on which is denoted as $\mathcal{L}_r$.
Similarly, $S_r$ can be represented as $S_r = \{ (q_r^i, D_r^i, ans_r^i) \mid i = 1, 2, \ldots, m_r \}$.

\subsection{Key Parameter Mining}
\label{sec: Importance-aware parameter unit type recognition}
The original pre-trained parameters across different layers store and process information in distinct ways, which influences their functionality and the types of tasks they are best suited for~\cite{chuang2023dola,meng2022locating}.
Therefore, we propose using the layer-wise activation levels of the original pre-trained parameters across various inputs to inform traditional gradient-based sensitivity calculations, in order to assess the overall contributions of different parameters to adherence and robustness, as follows: 
\paragraph{Forward activation probability computation.}
In the Transformer architecture, the FFN layer functions as a key-value unit capable of storing knowledge~\cite{meng2022locating} and is closely tied to the model's responses to different inputs~\cite{gurnee2024universal}, which can be formalized as follows:
\begin{equation}
\begin{aligned}
h^j=\left(\operatorname{Swish}(\tilde{h}^j W_1^j) \otimes \tilde{h}^j V^j\right) \cdot W_2^j,    
\end{aligned}
\end{equation}
where $W_1^j \in \mathbb{R}^{d \times 4 d}$, $V^j \in \mathbb{R}^{d \times 4 d}$, and $\boldsymbol{W}_2^j \in \mathbb{R}^{4 d \times d}$ are parameters, $\tilde{h}^j \in \mathbb{R}^d$ and $h^j \in \mathbb{R}^d$ represent the hidden states of a specific token at the $j$-th layer ($j \in\{1, \ldots, l\}$) after processing through MHA and FFN, respectively.
$d$ is the dimension of the hidden states.
SwiGLU is an activation function that has been widely adopted in recent LLM architectures~\cite{touvron2023llama,bai2023qwen}. In SwiGLU, the positive activation output of the Swish function indicates that the corresponding neurons are activated, thereby influencing the final prediction outcome through forward propagation~\cite{nair2010rectified}.
Inspired by~\citet{tang2024language}, we evaluate the sensitivity of the original pre-trained parameters to different abilities by calculating the expected positive activation outputs of neurons when faced with different inputs, i.e., $C_r = \left\{\left(q_r^i, D_r^i\right) \mid\left(q_r^i, D_r^i, a n s_r^i\right) \in S_r\right\}$ and $C_a = \left\{\left(q_a^i, D_a^i\right) \mid\left(q_a^i, D_a^i, a n s_a^i\right) \in S_a\right\}$: 
\begin{equation}
\small
\begin{aligned}
p_{*,k}^j = \mathbb{E} \left( \mathbb{I}\Bigl( \operatorname{Swish}(\tilde{h}^j W_1^j)_k > 0 \Bigl) \mid C_* \right),
\label{eq:single_activation}
\end{aligned}
\end{equation}
where $\operatorname{Swish}\left(\tilde{h}^j W_1^j\right)_k$ represents the activation value of the \( k \)-th neuron in the \( j \)-th layer. \( \mathbb{I} \) is the indicator function, and the \textbf{\( * \) denotes either adherence (symbolized by \( a \)) or robustness (symbolized by \( r \))}.
Next, we calculate the average activation probability \( \hat{p}_*^j \) of all FFN neurons within a single layer to measure the sensitivity of the original pre-trained parameters across different layers concerning adherence and robustness:
\begin{equation}
\small
\begin{aligned}
p_*^j= \frac{1}{4 d} \sum_{k=1}^{4 d} p_{*,k}^j,  
\quad 
\hat{p}_*^j =\frac{\exp \left(p_*^j\right)}{\sum_{j=1}^l \exp \left(p_*^j\right)}.
\end{aligned}
\label{eq:average_activation}
\end{equation}

\paragraph{Gradient-based sensitivity and uncertainty calculation.}
To quantify the sensitivity of parameters to training loss, we calculate the product of gradients and weights for all trainable parameters~\cite{molchanov2019importance,zhang2022platon}: 
\begin{equation}
I_*(w_{uv}) = |w_{uv} \cdot \nabla_{w_{uv}} \mathcal{L}_*|.
\label{eq:parameter_importance}
\end{equation}
This metric approximates the impact on the loss when a specific parameter is set to zero~\cite{molchanov2019importance}. However, it tends to exhibit high variance due to random sampling and the complex dynamics of the training process.
To mitigate this issue, we consider integrating sensitivity smoothing and uncertainty quantification~\cite{zhang2023adalora}: 
\begin{equation}
\small
\begin{aligned}
\overline{I}_*^{(t)}(w_{uv}) =&  \alpha_1 \overline{I}_*^{(t-1)}(w_{uv}) + (1 - \alpha_1) I_*^{(t)}(w_{uv}), \\
\overline{U}_*^{(t)}(w_{uv}) =&  \alpha_2 \overline{U}_*^{(t-1)}(w_{uv}) + (1 - \alpha_2) \\
& \cdot |I_*^{(t)}(w_{uv}) - \overline{I}_*^{(t)}(w_{uv})|,
\label{eq:ema_uncertainty}
\end{aligned}
\end{equation}
where $0<\alpha_1, \alpha_2<1$ are smoothing factors, and $t$ is the iteration number.

\paragraph{Importance score calculation and aggregation.}
Subsequently, we use \( \hat{p}_*^j \), obtained from the forward propagation, as a layer-specific clue to guide the sensitivity computation based on the gradient trajectory. 
Inspired by the computational approach in~\citet{zhang2023adalora}, we introduce a multiplicative operation that combines the activation levels of the original pre-trained parameters under different inputs with the smoothed sensitivity and uncertainty.
The final importance score for each parameter in terms of adherence or robustness (\(* \in \left\{a,r\right\}\)) is computed as follows:
   \begin{equation}
    \small
   \begin{aligned}
   \overline{s}_*^{(t)}(w_{uv}) = \hat{p}_*^j \cdot \overline{I}_*^{(t)}(w_{uv}) \cdot \overline{U}_*^{(t)}(w_{uv}),
   \label{eq:final_importance}
   \end{aligned}
   \end{equation}
where \( j \) represents the layer containing the parameter \( w_{uv} \). 
To assess the overall contribution of each parameter unit to model performance, we calculate the average importance of each individual parameter within the parameter unit:
\begin{equation}
\small
\begin{aligned}
    \mathcal{I}_*(e)=\frac{1}{d_1 \times d_2} \sum_{u=1}^{d_1} \sum_{v=1}^{d_2} \overline{s}_*\left(w_{u v}\right),
    \label{eq:unit_importance}
\end{aligned}
\end{equation}
where \(\mathcal{I}_*(e)\) measures the overall importance of all parameters within each parameter unit, with higher values indicating greater importance.
As a result, for model $\Theta$, which consists of \(n\) parameter units, we can obtain the distribution of importance scores $\mathcal{I}_*(\mathcal{E}) \in \mathbb{R}^n$ for adherence or robustness.

\subsection{Subspace Localization}
We then conduct an interactive analysis of importance distribution across different behaviors, decoupling and identifying various types of subspaces. 
We employ the Z-score, a common statistical measure~\cite{altman2017financial}, to standardize importance scores for adherence and robustness: 
\begin{equation}
\mathcal{Z}_*(e_i) = \Bigl({\mathcal{I}_*(e_i) - \mu_{*}} \Bigl) / {\sigma_{*}},
\label{eq:z_score_calculation}
\end{equation}
where \( \mathcal{Z}_*(e_i) \) is the standardized importance score of the \( i \)-th parameter unit for adherence or robustness, \( \mu_{*} \) is the mean score, and \( \sigma_{*} \) is the standard deviation. Based on $\left\{\mathcal{Z}_*\left(e_i\right) \mid i \in \{1, \ldots, n\}\right\}$, we identify and locate four types of subspaces:

\paragraph{1. Entangled Subspace.}
This subspace consists of parameter units that are crucial for both adherence and robustness, and we assume it represents the model's ability to perceive and analyze context (queries and external knowledge).
Specifically, we select parameter units with Z-scores greater than 1 for both adherence and robustness: 
\begin{equation}
\small
    \mathcal{E}_c=\left\{e_i \mid \mathcal{Z}_a\left(e_i\right)>1 \ \land \ \mathcal{Z}_r\left(e_i\right)>1; i \in \{1, \ldots, n\}
    \right\}.
\label{eq:z_score_intersecting}
\end{equation}

\paragraph{2. Adherence Subspace.}
This subspace is composed of parameter units that are important for adherence but not for robustness. It relates to the model's ability to solve problems by replicating and summarizing contextual information: 
     \begin{equation}
    \small
    \mathcal{E}_{ax}=\left\{e_i \mid \mathcal{Z}_a\left(e_i\right)>1  \ \land \ \mathcal{Z}_r\left(e_i\right)<1; i \in \{1, \ldots, n\}\right\}.
     \label{eq:z_score_adherence}
     \end{equation}

\paragraph{3. Robustness Subspace.}
This subspace comprises parameter units that are crucial for robustness but less significant for adherence. It is associated with the model's capacity to solve problems by retrieving internal memories: 
     \begin{equation}
    \small
    \mathcal{E}_{rx}=\left\{e_i \mid \mathcal{Z}_r\left(e_i\right)>1   \ \land \ \mathcal{Z}_a\left(e_i\right)<1; i \in \{1, \ldots, n\}\right\}.
     \label{eq:z_score_robustness}
     \end{equation}

\paragraph{4. Other Subspaces.}
These subspaces consist of all remaining parameter units outside the three previously mentioned subspaces, representing the other capabilities of LLMs, denoted as $\mathcal{E}_{o}$.

\subsection{Type-Tailored Tuning}
After decoupling and identifying key subspaces related to knowledge control, we develope specific and appropriate fine-tuning strategies tailored to each subspace.

\subsubsection{Document Extraction Task}
\label{section:evidence}

To more effectively optimize the entangled subspace, we design a document extraction dataset \( S_c = \{ (q_c^i, D_c^i, ans_c^i) \mid i = 1, 2, \ldots, m_c \} \).  Specifically, we curate documents from the SQuAD2.0 dataset, gathering three types of documents for each question: relevant documents that contain the question's answer, noise documents that are on the same topic, and noise documents from different topics, simulating a variety of retrieval context scenarios.
In this document extraction task, we present the mixed context of the question along with the three types of documents as input and expect the LLMs to identify the specific document types and accurately restate their content. We denote the cross-entropy loss on \( S_c \) as \( \mathcal{L}_c \).
To simultaneously enhance adherence and robustness, we train $\mathcal{E}_c$ using a combination of $S_a$, $S_r$, and $S_c$:
\begin{equation}
\small
\begin{aligned}
&\mathcal{Z}_*\left(\mathcal{E}_c\right)=\mathbb{E}\left[\mathcal{Z}_*\left(e_i\right) \mid e_i \in \mathcal{E}_c\right], \\
&\gamma_*=\frac{\exp \left(\mathcal{Z}_*\left(\mathcal{E}_c\right)\right)}{\exp \left(\mathcal{Z}_a\left(\mathcal{E}_c\right)\right)+\exp \left(\mathcal{Z}_r\left(\mathcal{E}_c\right)\right)}, \\
&\mathcal{L}_{c x}=\delta_1\left(\gamma_a \times \mathcal{L}_a+\gamma_r \times \mathcal{L}_r\right)+\left(1-\delta_1\right) \mathcal{L}_c,
\end{aligned}
\label{eq: loss-c}
\end{equation}
where $0 < \delta_1 < 1$ acts as a reweighting factor between the original task and the added task. We determine the weights of the two loss terms adaptively by calculating expected Z-scores for adherence and robustness on \(\mathcal{E}_c\).

\subsubsection{Boundary-Controlled Fine-Tuning}
To prevent contradictory supervision signals from contaminating the adherence and robustness subspaces, we propose a boundary-controlled fine-tuning strategy:

\begin{itemize}[leftmargin=*]
\item \textbf{Adherence Subspace.} For $\mathcal{E}_{ax}$, we ensure that it remains unaffected by gradients related to robustness, allowing the LLMs' adherence capabilities to be fully optimized: 
   \begin{equation}
   \mathcal{L}_{ax} = \delta_1 \mathcal{L}_{a} + (1-\delta_1) \mathcal{L}_{c}.
    \label{eq: loss-ax}
   \end{equation}
\item \textbf{Robustness Subspace.} For $\mathcal{E}_{rx}$, we ensure that it remains unaffected by gradients related to adherence, thereby enabling a breakthrough in enhancing robustness:  
   \begin{equation}
   \mathcal{L}_{rx} = \delta_1 \mathcal{L}_{r} + (1-\delta_1) \mathcal{L}_{c}.
    \label{eq: loss-rx}
   \end{equation}
\end{itemize}

Additionally, we maintain the initialized weights of $\mathcal{E}_{o}$ to prevent divergence from pre-trained weights, preserving the other capabilities of LLMs.
We present the detailed algorithm of \mname in Appendix~\ref{appendix:algorithm}.

\section{Experiment}
\subsection{Experimental Setup}
\paragraph{Datasets.} 
We construct the \mname training dataset based on SQuAD2.0~\cite{rajpurkar2018knowdontknowunanswerable}. The test datasets comprise the following three types: 
(1) \textbf{SQuAD2.0-Eval}. Following~\citet{li2023large}, we further simulate complex retrieval scenarios using annotation information from SQuAD2.0 to more effectively evaluate the adherence and robustness of RALMs, resulting in the creation of SQuAD2.0-Eval.
(2) \textbf{Open-source RAG datasets:} RGB~\cite{chen2024benchmarking} and KNOT~\cite{liu2024untangle} are two general-domain QA datasets designed to evaluate LLMs' ability to effectively utilize retrieved information while withstanding various imperfections in the retrieval process.
(3) \textbf{Domain-specific dataset:} CMB~\cite{wang2024cmb} is a multi-task QA dataset in the medical domain, comprising 269,359 questions across various categories and disciplines. Due to quantity constraints, we randomly sample 4,000 questions for testing. Further details about the datasets are provided in the Appendix~\ref{appedix:data_details}.

\paragraph{Baselines.} 
We evaluate \mname against the following widely-used and state-of-the-art RAG baseline methods:
\textbf{\emph{Base Model (Base)}}
responds to queries by leveraging external knowledge, functioning as the fundamental retrieve-read framework~\cite{gao2023retrieval}. We choose LLaMA2-7B-Chat~\cite{touvron2023llama} and Qwen1.5-14B-Chat~\cite{team2024introducing} as the base models and investigate the improvements introduced by \mname.
\textbf{\emph{Prompt-based Method (Prompt)}}
~\cite{zhou2023context} utilizes a carefully designed prompt strategy to enhance adherence.
\textbf{\emph{COT-VE}}
~\cite{zhao2023verify} guides LLMs to edit potentially incorrect or outdated rationales using external knowledge.
\textbf{\emph{KAFT}}
~\cite{li2023large} enhances the adherence and robustness of LLMs by constructing a specific fine-tuning dataset.
\textbf{\emph{CAD}}
~\cite{shi2024trusting} reduces the model's reliance on internal knowledge through contrastive decoding.
\textbf{\emph{RAAT}}
~\cite{fang2024enhancing} improves the noise robustness of LLMs through adaptive adversarial training.
\textbf{\emph{IRCAN}}
~\cite{shi2024ircan} enhances the adherence of LLMs by detecting and boosting adherence-related neurons.

It is worth noting that the baseline method KAFT is implemented by training all subspaces using the full training dataset uniformly. 
To ensure the fairness of experimental comparisons, for the training-based baseline methods KAFT and RAAT, we construct training datasets based on the same source data as \mname, following the strategies outlined in their original methods.
As for the IRCAN method, which relies on a specific dataset to identify context-aware neurons, we likewise use the same dataset to conduct the corresponding neuron detection task.

We also compare \mname with three reduced variants to conduct ablation studies:
\begin{itemize}[leftmargin=*,noitemsep,topsep=3pt]
\item $\text{\mname}_{l-}$ removes layer-level clues based on forward activations when mining key parameters.
\item $\text{\mname}_{b-}$ trains $\mathcal{E}_c$, $\mathcal{E}_{ax}$, and $\mathcal{E}_{rx}$ using three datasets (\( S_a \), \( S_r \), and \( S_c \)) simultaneously.
\item $\text{\mname}_{e-}$ removes the document extraction task during type-tailored tuning.
\end{itemize}

\paragraph{Metrics.}
In line with~\citet{li2023large}, we evaluate the adherence and robustness of RALMs using two distinct metrics.
For adherence, we supplement LLMs with conflicting contexts from the test set as additional knowledge and measure the proportion of cases where the LLMs follow the conflicting evidence to generate their responses, denoted as $R_{Ad}$.
Regarding robustness, we use irrelevant contexts from the test set as supplementary knowledge and measure the proportion of cases where the model successfully avoids extracting answers from these irrelevant contexts, which is denoted as $R_{Ro}$.
Higher values of $R_{Ad}$ and $R_{Ro}$ indicate better adherence and robustness, respectively.

More implementation details and experimental results are provided in Appendix~\ref{appendix:imple},  \ref{appendix:general_performance}, \ref{appendix:sensitivity}, \ref{appendix:memorize}, \ref{appendix:additional_vis}, \ref{appendix:diverse_evaluations}, \ref{appendix:computation_cost}, and \ref{appendix:case_study}.

\begin{table*}[!t]
\setlength{\tabcolsep}{1mm}
\fontsize{9pt}{11pt}\selectfont   
\centering
\resizebox{1.0\linewidth}{!}{
\begin{tabular}{cc|cc|cc|cc|cc|cc|cc}
\hline
\multirow{3}{*}{\textbf{Method}} & \multicolumn{1}{c|}{\textbf{LLM}} & \multicolumn{6}{c|}{LLaMA2-7B-Chat} & \multicolumn{6}{c}{Qwen1.5-14B-Chat} \\
\cline{2-14}
& \multicolumn{1}{c|}{\textbf{Dataset}} & \multicolumn{2}{c|}{SQuAD} & \multicolumn{2}{c|}{RGB} & \multicolumn{2}{c|}{KNOT} & \multicolumn{2}{c|}{SQuAD} & \multicolumn{2}{c|}{RGB} & \multicolumn{2}{c}{KNOT} \\
\cline{2-14}
& \textbf{Metric} & $R_{Ad}$ & $R_{Ro}$ & $R_{Ad}$ & $R_{Ro}$ &$R_{Ad}$ & $R_{Ro}$ &$R_{Ad}$ & $R_{Ro}$ &$R_{Ad}$ & $R_{Ro}$ & $R_{Ad}$ & $R_{Ro}$   \\
\hline
\multirow{8}{*}{\textbf{Baselines}} & Base &{44.20}&{16.40}&{68.00}&{29.50}&{45.09}&{20.54}&{59.71}&{21.37}&{68.50}&{34.00}&{52.73}&{22.69} \\
& Prompt &{46.40}&{9.50}&{69.50}&{17.50}&{43.74}&{15.01}&{63.69}&{18.29}&{78.00}&{32.50}&{53.19}&{16.77} \\
& COT-VE &{47.45}&{17.20}&{68.50}&{31.00}&{46.39}&{21.27}&{64.23}&{25.94}&{66.00}&{38.00}&{55.42}&{24.59} \\
& KAFT & \underline{54.15}&{18.43}& {71.50}&{30.50}& \underline{47.09}&{22.92}& {75.55}&{28.00}& \underline{82.50}&{39.50}& \underline{56.23}&{25.94} \\
& CAD &{45.72}&{8.80}&{70.00}&{16.00}&{44.14}&{12.79}&{64.98}&{17.20}&{72.00}&{28.50}&{53.14}&{14.79} \\
& RAAT &{39.25}& \underline{40.73}&{49.50}& \underline{41.00}&{25.09}& \underline{35.58}&{60.30}& \underline{40.61}&{71.50}& \underline{42.50}&{46.87}& \underline{37.92} \\
& IRCAN & {53.17}&{13.50}&\underline{72.50}&{20.00}&{46.51}&{16.50}&\underline{76.24}&{16.15}&{82.00}&{30.50}&{55.21}&{17.45} \\
\cdashline{1-14}
\textbf{Ours} & \textbf{\mname} &\textbf{69.24}&\textbf{44.85}&\textbf{79.50}&\textbf{45.50}&\textbf{67.42}&\textbf{42.82}&\textbf{80.47}&\textbf{47.58}&\textbf{84.50}&\textbf{46.50}&\textbf{73.27}&\textbf{43.34} \\
\cdashline{1-14}
\multirow{3}{*}{\textbf{Ablation}} & $\text{\mname}_{l-}$ & {66.15}& {39.78}& {77.00}& {40.00}& {64.35}& {37.63}& {78.06}& {42.25}& {83.50}& {42.50}& {67.87}& {40.91} \\
& $\text{\mname}_{b-}$ &{55.90}&{20.70}&{73.50}&{31.50}&{49.26}&{23.05}&{77.28}&{30.92}&{83.00}&{41.50}&{58.73}&{27.36} \\
& $\text{\mname}_{e-}$ &{62.57}&{36.71}&{75.50}&{37.50}&{60.55}&{34.98}&{77.59}&{38.24}&{83.00}& {42.50}&{64.13}&{37.28} \\

\hline
\end{tabular}}
\caption{\label{tab:comparison}Performance comparisons (\%) on SQuAD2.0-Eval (SQuAD), RGB, and KNOT. The best performance is in \textbf{boldface} and the best results among the baselines are \underline{underlined}. }
\end{table*}

\subsection{Experimental Results}
\subsubsection{Main Results}

Table~\ref{tab:comparison} displays the performance of \mname compared to baseline methods across three datasets. Overall, \textbf{our proposed \mname excels on all backbone models and datasets, significantly surpassing current state-of-the-art methods.} This highlights the effectiveness of \mname in fine-grained decoupling in parameter space and its customized fine-tuning capabilities.
From Table~\ref{tab:comparison}, we can also observe several insights. Firstly, methods that rely on LLMs' basic interaction capabilities at the prompt level, such as Prompt and CoT-VE, often depend indiscriminately on external knowledge due to LLMs' limited noise recognition capabilities. Secondly, IRCAN and RAAT emphasize adherence and robustness respectively, and each inevitably sacrifices another key capability when integrating knowledge.
Additionally, a notable phenomenon is that, compared to the baseline, \textbf{\mname achieves a more balanced improvement in both adherence and robustness.} 
This is because we avoid contaminating key parameters with conflicting supervisory signals during the fine-tuning process.

\subsubsection{Ablation Study}
To fully understand the contribution of each component in \mname to the overall performance, we conducted ablation studies. As shown in Table~\ref{tab:comparison}, the performance decline of $\text{\mname}_{l-}$ compared to \mname indicates that the hierarchical clues provided by forward activations are essential for precisely locating parameters closely related to adherence and robustness. The superior performance of \mname over $\text{\mname}_{b-}$ and $\text{\mname}_{e-}$ demonstrates that our proposed type-tailored tuning strategy effectively mitigates the negative impact of conflicting supervision signals on the model’s behavior, allowing for the full optimization of both adherence and robustness.

\subsection{Analysis}

\subsubsection{Generalization Analysis}
\label{sec:generalization_analysis}
\begin{table}[t]
\centering
\begin{tabular}{lcccc}
\toprule
\multirow{2}{*}{Method} & \multicolumn{2}{c}{LLaMA2-7B} & \multicolumn{2}{c}{Qwen1.5-14B} \\
\cmidrule(lr){2-5}
                     & $R_{Ad}$          & $R_{Ro}$         & $R_{Ad}$         & $R_{Ro}$         \\
\midrule
Base      &    54.28      &  20.17       &    59.07        &           21.39 \\
\mname    &     \textbf{75.79}        &         \textbf{48.21}    &      \textbf{79.63}     &      \textbf{45.55}     \\ 
\bottomrule
\end{tabular}
\caption{\label{table:generalize}Performance comparisons (\%) on CMB.}
\end{table}

To demonstrate the strong generalization capability of \mname beyond general-domain training sets, we conduct experiments on the medical benchmark CMB~\cite{wang2024cmb}. By utilizing medical triples and documents from the Huatuo-26M dataset~\cite{li2023huatuo} to build supplementary medical context, we create a dataset to evaluate the medical knowledge utilization of LLMs, based on 4,000 CMB questions.
As shown in Table~\ref{table:generalize}, LLMs trained with \mname in a general domain continue to demonstrate superior knowledge selection capabilities when transferred to specific vertical domains. These results further \textbf{validate the generalization ability and effectiveness of \mname.}

\subsubsection{Behavioral Tendency Exploration}
\begin{figure}[t]
  \centering
  \includegraphics[scale=0.31]{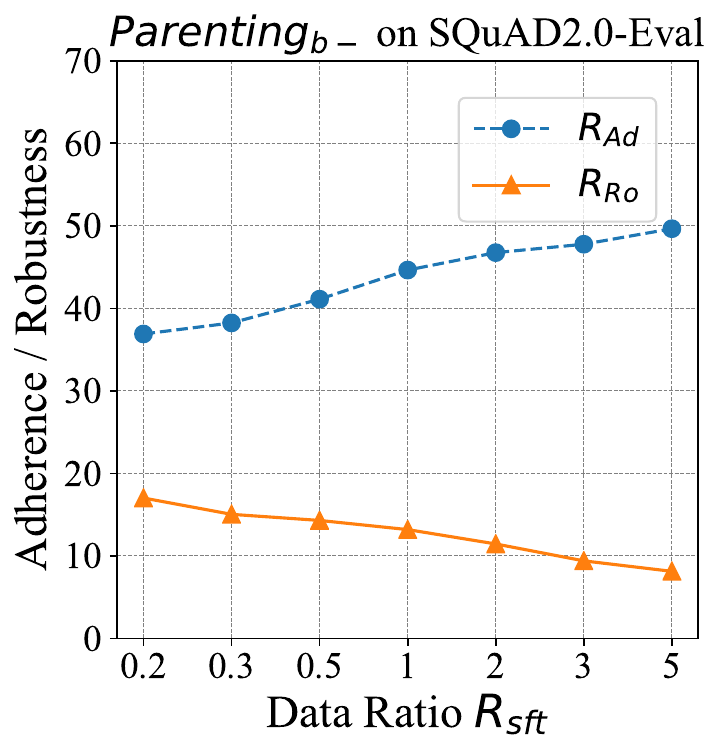}
  \includegraphics[scale=0.31]{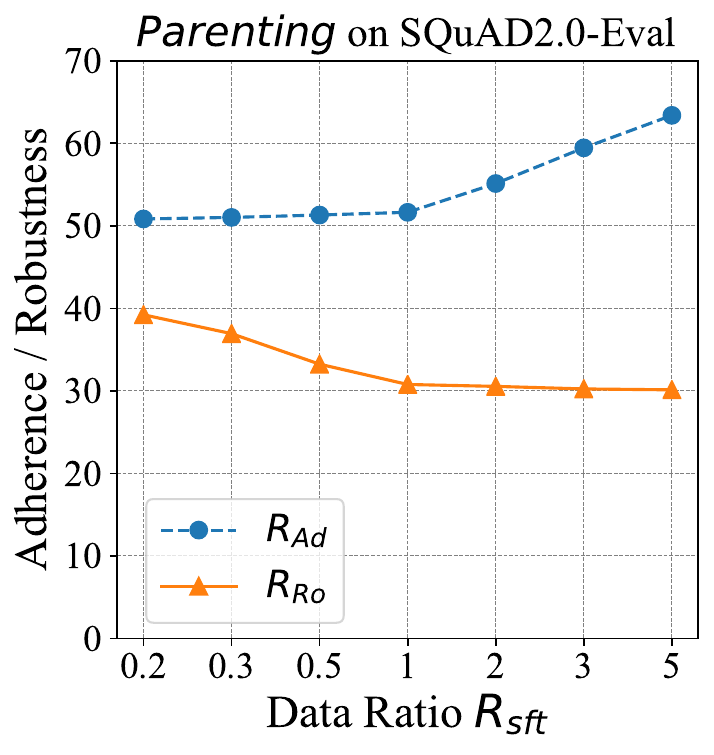}
  \caption{
  Analysis of the behavioral patterns of \mname (\textbf{Right}) and $\text{\mname}_{b-}$ (\textbf{Left}) under varying data ratios (on SQuAD2.0-Eval with LLaMA2-7B-Chat).
  }
   \label{fig:data_ratio}
\end{figure}

To gain a more intuitive understanding of \mname's ability to balance adherence and robustness, we explore the behavioral patterns of \mname and $\text{\mname}_{b-}$ under different ratios of supervised data.
Specifically, for the two types of supervision signals that are challenging to balance due to their contradictory nature, we randomly select two training subsets, \( S_a^{sub} \) and \( S_r^{sub} \), from \( S_a \) and \( S_r \) respectively, with \( |S_a^{sub}| = |S_r^{sub}| = |S_a| / 2 \).
Within these subsets, we continuously adjust the ratio between these two signals, denoted as \( R = |S_a^{sub}| / |S_r^{sub}| \). When \( R>1 \), we keep \( |S_r^{sub}| \) constant, and when \( R<1 \), we keep \( |S_a^{sub}| \) constant.
As shown in Figure~\ref{fig:data_ratio}, as \( R \) increases, the adherence of the LLM fine-tuned with \mname steadily rises, while its robustness remains stable. Conversely, as \( R \) decreases, the robustness of the LLMs improves, with adherence remaining largely unchanged. However, in the case of $\text{\mname}_{b-}$, there is a tug-of-war between adherence and robustness, where an increase in the proportion of one signal results in a significant decline in the other, preventing full optimization of both capabilities. 
This indicates that separating the conflicting supervision signals allows for a more balanced optimization of adherence and robustness.

\subsubsection{Visualization of Parameter Units}
\label{section:visualization_llama2}
\begin{figure}[t]
  \centering
    \includegraphics[width=1\linewidth]{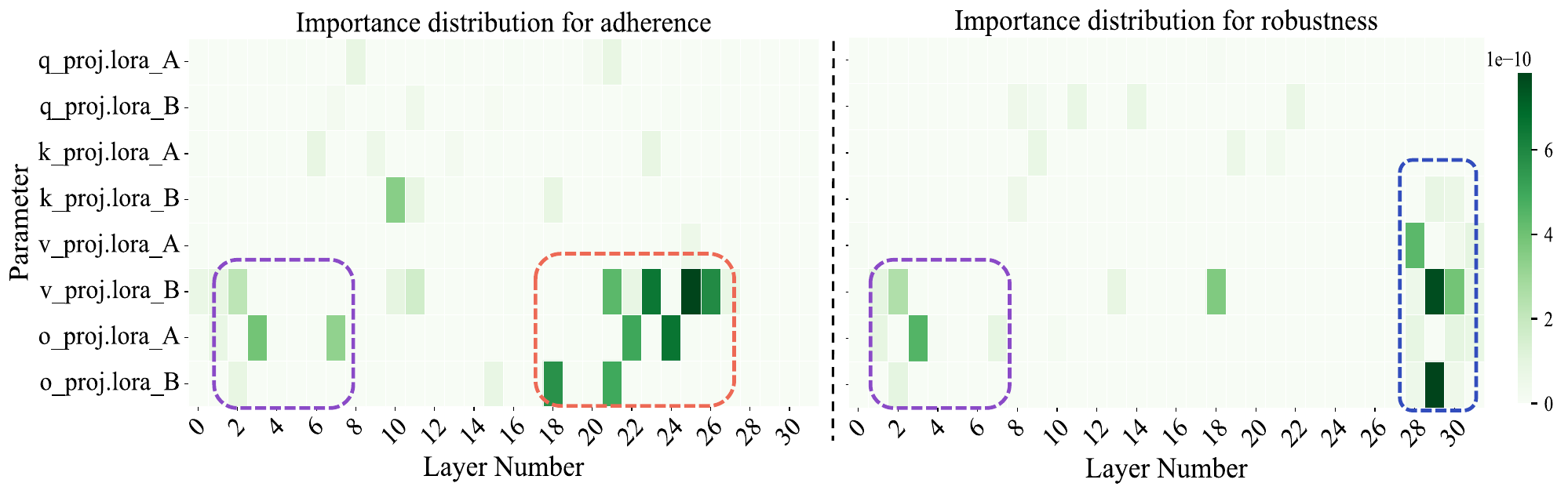}
  \caption{Visualization of parameter importance distributions $\mathcal{I}_a(\mathcal{E})$ and $\mathcal{I}_r(\mathcal{E})$ for adherence (\textbf{Left}) and robustness (\textbf{Right}) in LLaMA2-7B-Chat.
  }
  \label{fig:heatmap-llama2}
\end{figure}

Figure~\ref{fig:heatmap-llama2} visualizes the distribution of parameter unit importance for adherence and robustness in LLaMA2-7B-Chat. We can observe the parameter units (entangled) within the Entangled Subspace (purple boxes). It also highlights the parameter units (adherence-specific) in the Adherence Subspace (the red box) and those (robustness-specific) in the Robustness Subspace (the blue box).
Additionally, we observe that adherence-specific units are primarily located in the middle and upper-middle layers. This aligns with findings from studies~\citet{chuang2023dola} and~\citet{dai2022knowledge}, which indicate that these layers play a significant role when LLMs copy information from inputs. 
Robustness-specific units are mainly found in the upper layers, with a few in the middle layers. 
This is consistent with the study~\citet{chuang2023dola}, suggesting that internal factual knowledge is typically encoded in the higher layers of LLMs. 
We also note that entangled parameter units are predominantly located in the middle to lower-middle layers, aligning with the study~\citet{fan2024not}.

\subsection{Noise Recognition Capability}
\label{appendix:noise_rec}
\begin{table}
\centering
\scalebox{0.9}{
\begin{tabular}{lcc}
\toprule
Method & SQuAD & RGB\\
\midrule
\rule{0pt}{6pt} LLaMA2-7B-Chat & 55.49 & 63.50 \\
\midrule
\rule{0pt}{8pt} KAFT & 59.24 & 65.00 \\
\rule{0pt}{8pt} RAAT & 62.48 & 67.00 \\
\rule{0pt}{8pt} \mname & \textbf{69.89}  & \textbf{72.50} \\
\bottomrule
\end{tabular}}
\caption{\label{table:noise_regonition}Performance comparisons of \mname with other RAG methods (all utilizing the same LLaMA2-7B-Chat backbone) on the noise identification task. All results are reported in accuracy (\%).}
\end{table}

To further validate the effectiveness of our framework in enhancing the context-awareness of LLMs, we construct a noise identification experiment on the SQuAD2.0-Eval and RGB datasets to evaluate the model's ability to analyze the association between queries and external knowledge.
In detail, for each question, we select a topic-relevant document based on the dataset’s annotation information. This document may either provide supporting evidence or serve as a noise document that does not answer the question. 
We then prompt LLMs to classify the type of document. 
The dataset contains a balanced number of labels for both evidence and noise categories, and accuracy is used as the evaluation metric for this task.
We focus on the comparison between \mname and two existing frameworks: KAFT and RAAT. The results in Table~\ref{table:noise_regonition} show that \mname stands out among the evaluated frameworks, particularly surpassing RAAT, which is specifically designed to enhance noise robustness. 
This result highlights the effectiveness of \mname in enhancing the context-awareness capabilities of LLMs.

\section{Conclusion}
In this work, we propose a novel and versatile RAG knowledge integration framework, dubbed \mname, which optimizes LLMs' knowledge selection through parameter decoupling and tailored tuning, thereby establishing an effective control mechanism for both internal and external knowledge.
\mname first performs a fine-grained analysis and decoupling of adherence and robustness in the parameter space using a key parameter mining method. It then applies a type-tailored tuning strategy to systematically and thorough optimize behavior-specific parameters. 
Experimental results across multiple datasets and models consistently demonstrate the effectiveness of \mname.

\section*{Limitations}
Despite the promising results obtained in our work, it is important to acknowledge its limitations.

On the one hand, we inevitably encountered some errors where the model failed to adhere to conflicting external knowledge, particularly on the KNOT dataset, which requires knowledge-based reasoning. 
According to the original setup of the dataset~\cite{liu2024untangle}, 
we observed that the vast majority of these errors fell under the KNOT-I category, which requires implicit reasoning: the model must independently decompose the question, infer a plausible reasoning path, and derive the answer based on conflicting knowledge.
We believe this is primarily due to the current limitations of LLMs in actively integrating internal and external knowledge for reasoning. In the future, we plan to explore the inclusion of more diverse datasets and investigate finer-grained parameter unit partitioning to further enhance the model's reasoning capabilities.
First, within the \mname framework, we treat an individual matrix in LLMs as a parameter unit, yet this granularity may still introduce redundancy within the matrix. There could be a possibility to further refine the model's capability segmentation, such as decomposing the product of two low-rank matrices in LoRA into even finer units~\cite{wang2023orthogonal,feng2024kif}, which might enhance model performance. Additionally, the construction of our training dataset solely relied on resources from SQuAD2.0. Although we have demonstrated the generalized capabilities of the \mname fine-tuned model across multiple benchmark datasets, there is still room for improvement. In the future, we plan to gather a more diverse array of high-quality datasets, aim to launch a more effective version of \mname, and establish a broader and more varied benchmark to assess the knowledge integration abilities of RALMs.

On the other hand, although we focus on the reader phase (where LLMs utilize knowledge to answer questions) and enhance model robustness through instruction tuning, which has been empirically validated as effective, the overall performance of RAG systems remains constrained by interference from low-quality documents. This issue is particularly pronounced when content is sourced from external platforms such as the web, where poor-quality materials can significantly degrade generation quality. Addressing this challenge remains non-trivial. To mitigate this issue, we plan to explore training a discriminator capable of identifying low-quality documents and incorporating its results into the prompt. Additionally, we will consider calibrating the uncertainty of LLMs and using it as a clue for evaluating document quality.

\section*{Acknowledgments}
This work is supported by the National Natural Science Foundation of China (No.62172011).

\bibliography{custom}

\begin{thebibliography}{84}
\providecommand{\natexlab}[1]{#1}

\bibitem[{Achiam et~al.(2023)Achiam, Adler, Agarwal, Ahmad, Akkaya, Aleman, Almeida, Altenschmidt, Altman, Anadkat et~al.}]{achiam2023gpt}
Josh Achiam, Steven Adler, Sandhini Agarwal, Lama Ahmad, Ilge Akkaya, Florencia~Leoni Aleman, Diogo Almeida, Janko Altenschmidt, Sam Altman, Shyamal Anadkat, et~al. 2023.
\newblock Gpt-4 technical report.
\newblock \emph{arXiv preprint arXiv:2303.08774}.

\bibitem[{Altman et~al.(2017)Altman, Iwanicz-Drozdowska, Laitinen, and Suvas}]{altman2017financial}
Edward~I Altman, Ma{\l}gorzata Iwanicz-Drozdowska, Erkki~K Laitinen, and Arto Suvas. 2017.
\newblock Financial distress prediction in an international context: A review and empirical analysis of altman's z-score model.
\newblock \emph{Journal of international financial management \& accounting}, 28(2):131--171.

\bibitem[{Asai et~al.()Asai, Wu, Wang, Sil, and Hajishirzi}]{asaiself}
Akari Asai, Zeqiu Wu, Yizhong Wang, Avirup Sil, and Hannaneh Hajishirzi.
\newblock Self-rag: Learning to retrieve, generate, and critique through self-reflection.
\newblock In \emph{The Twelfth International Conference on Learning Representations}.

\bibitem[{Bai et~al.(2023)Bai, Bai, Chu, Cui, Dang, Deng, Fan, Ge, Han, Huang et~al.}]{bai2023qwen}
Jinze Bai, Shuai Bai, Yunfei Chu, Zeyu Cui, Kai Dang, Xiaodong Deng, Yang Fan, Wenbin Ge, Yu~Han, Fei Huang, et~al. 2023.
\newblock Qwen technical report.
\newblock \emph{arXiv preprint arXiv:2309.16609}.

\bibitem[{Brown(2020)}]{brown2020language}
Tom~B Brown. 2020.
\newblock Language models are few-shot learners.
\newblock \emph{arXiv preprint arXiv:2005.14165}.

\bibitem[{Bubeck et~al.(2023)Bubeck, Chandrasekaran, Eldan, Gehrke, Horvitz, Kamar, Lee, Lee, Li, Lundberg et~al.}]{bubeck2023sparks}
S{\'e}bastien Bubeck, Varun Chandrasekaran, Ronen Eldan, Johannes Gehrke, Eric Horvitz, Ece Kamar, Peter Lee, Yin~Tat Lee, Yuanzhi Li, Scott Lundberg, et~al. 2023.
\newblock Sparks of artificial general intelligence: Early experiments with gpt-4.
\newblock \emph{arXiv preprint arXiv:2303.12712}.

\bibitem[{Chen et~al.(2024)Chen, Lin, Han, and Sun}]{chen2024benchmarking}
Jiawei Chen, Hongyu Lin, Xianpei Han, and Le~Sun. 2024.
\newblock Benchmarking large language models in retrieval-augmented generation.
\newblock In \emph{Proceedings of the AAAI Conference on Artificial Intelligence}, volume~38, pages 17754--17762.

\bibitem[{Chowdhery et~al.(2023)Chowdhery, Narang, Devlin, Bosma, Mishra, Roberts, Barham, Chung, Sutton, Gehrmann et~al.}]{chowdhery2023palm}
Aakanksha Chowdhery, Sharan Narang, Jacob Devlin, Maarten Bosma, Gaurav Mishra, Adam Roberts, Paul Barham, Hyung~Won Chung, Charles Sutton, Sebastian Gehrmann, et~al. 2023.
\newblock Palm: Scaling language modeling with pathways.
\newblock \emph{Journal of Machine Learning Research}, 24(240):1--113.

\bibitem[{Chuang et~al.(2023)Chuang, Xie, Luo, Kim, Glass, and He}]{chuang2023dola}
Yung-Sung Chuang, Yujia Xie, Hongyin Luo, Yoon Kim, James Glass, and Pengcheng He. 2023.
\newblock Dola: Decoding by contrasting layers improves factuality in large language models.
\newblock \emph{arXiv preprint arXiv:2309.03883}.

\bibitem[{Clark et~al.(2018)Clark, Cowhey, Etzioni, Khot, Sabharwal, Schoenick, and Tafjord}]{clark2018think}
Peter Clark, Isaac Cowhey, Oren Etzioni, Tushar Khot, Ashish Sabharwal, Carissa Schoenick, and Oyvind Tafjord. 2018.
\newblock Think you have solved question answering? try arc, the ai2 reasoning challenge.
\newblock \emph{arXiv preprint arXiv:1803.05457}.

\bibitem[{Cobbe et~al.(2021)Cobbe, Kosaraju, Bavarian, Chen, Jun, Kaiser, Plappert, Tworek, Hilton, Nakano et~al.}]{cobbe2021training}
Karl Cobbe, Vineet Kosaraju, Mohammad Bavarian, Mark Chen, Heewoo Jun, Lukasz Kaiser, Matthias Plappert, Jerry Tworek, Jacob Hilton, Reiichiro Nakano, et~al. 2021.
\newblock Training verifiers to solve math word problems.
\newblock \emph{arXiv preprint arXiv:2110.14168}.

\bibitem[{Creswell et~al.()Creswell, Shanahan, and Higgins}]{creswellselection}
Antonia Creswell, Murray Shanahan, and Irina Higgins.
\newblock Selection-inference: Exploiting large language models for interpretable logical reasoning.
\newblock In \emph{The Eleventh International Conference on Learning Representations}.

\bibitem[{Dai et~al.(2022)Dai, Dong, Hao, Sui, Chang, and Wei}]{dai2022knowledge}
Damai Dai, Li~Dong, Yaru Hao, Zhifang Sui, Baobao Chang, and Furu Wei. 2022.
\newblock Knowledge neurons in pretrained transformers.
\newblock In \emph{Proceedings of the 60th Annual Meeting of the Association for Computational Linguistics (Volume 1: Long Papers)}, pages 8493--8502.

\bibitem[{Fan et~al.(2024{\natexlab{a}})Fan, Jiang, Li, Meng, Han, Shang, Sun, Wang, and Wang}]{fan2024not}
Siqi Fan, Xin Jiang, Xiang Li, Xuying Meng, Peng Han, Shuo Shang, Aixin Sun, Yequan Wang, and Zhongyuan Wang. 2024{\natexlab{a}}.
\newblock Not all layers of llms are necessary during inference.
\newblock \emph{arXiv preprint arXiv:2403.02181}.

\bibitem[{Fan et~al.(2024{\natexlab{b}})Fan, Ding, Ning, Wang, Li, Yin, Chua, and Li}]{fan2024survey}
Wenqi Fan, Yujuan Ding, Liangbo Ning, Shijie Wang, Hengyun Li, Dawei Yin, Tat-Seng Chua, and Qing Li. 2024{\natexlab{b}}.
\newblock A survey on rag meeting llms: Towards retrieval-augmented large language models.
\newblock In \emph{Proceedings of the 30th ACM SIGKDD Conference on Knowledge Discovery and Data Mining}, pages 6491--6501.

\bibitem[{Fang et~al.(2024)Fang, Bai, Ni, Yang, Chen, and Xu}]{fang2024enhancing}
Feiteng Fang, Yuelin Bai, Shiwen Ni, Min Yang, Xiaojun Chen, and Ruifeng Xu. 2024.
\newblock Enhancing noise robustness of retrieval-augmented language models with adaptive adversarial training.
\newblock \emph{arXiv preprint arXiv:2405.20978}.

\bibitem[{Feng et~al.(2024{\natexlab{a}})Feng, Chu, Xu, Lu, Liu, Yu, and Wu}]{feng2024kif}
Yujie Feng, Xu~Chu, Yongxin Xu, Zexin Lu, Bo~Liu, Philip~S Yu, and Xiao-Ming Wu. 2024{\natexlab{a}}.
\newblock Kif: Knowledge identification and fusion for language model continual learning.
\newblock \emph{arXiv preprint arXiv:2408.05200}.

\bibitem[{Feng et~al.(2024{\natexlab{b}})Feng, Chu, Xu, Shi, Liu, and Wu}]{feng2024tasl}
Yujie Feng, Xu~Chu, Yongxin Xu, Guangyuan Shi, Bo~Liu, and Xiao-Ming Wu. 2024{\natexlab{b}}.
\newblock Tasl: Continual dialog state tracking via task skill localization and consolidation.
\newblock In \emph{Proceedings of the 62nd Annual Meeting of the Association for Computational Linguistics (Volume 1: Long Papers)}, pages 1266--1279.

\bibitem[{Feng et~al.(2023)Feng, Lu, Liu, Zhan, and Wu}]{feng2023towards}
Yujie Feng, Zexin Lu, Bo~Liu, Liming Zhan, and Xiao-Ming Wu. 2023.
\newblock Towards llm-driven dialogue state tracking.
\newblock In \emph{Proceedings of the 2023 Conference on Empirical Methods in Natural Language Processing}, pages 739--755.

\bibitem[{Feng et~al.(2025)Feng, Wang, Lu, Fu, Shi, Xu, Wang, Yu, Chu, and Wu}]{feng2025recurrent}
Yujie Feng, Xujia Wang, Zexin Lu, Shenghong Fu, Guangyuan Shi, Yongxin Xu, Yasha Wang, Philip~S Yu, Xu~Chu, and Xiao-Ming Wu. 2025.
\newblock Recurrent knowledge identification and fusion for language model continual learning.
\newblock \emph{arXiv preprint arXiv:2502.17510}.

\bibitem[{Gao et~al.(2023)Gao, Xiong, Gao, Jia, Pan, Bi, Dai, Sun, and Wang}]{gao2023retrieval}
Yunfan Gao, Yun Xiong, Xinyu Gao, Kangxiang Jia, Jinliu Pan, Yuxi Bi, Yi~Dai, Jiawei Sun, and Haofen Wang. 2023.
\newblock Retrieval-augmented generation for large language models: A survey.
\newblock \emph{arXiv preprint arXiv:2312.10997}.

\bibitem[{Gurnee et~al.(2024)Gurnee, Horsley, Guo, Kheirkhah, Sun, Hathaway, Nanda, and Bertsimas}]{gurnee2024universal}
Wes Gurnee, Theo Horsley, Zifan~Carl Guo, Tara~Rezaei Kheirkhah, Qinyi Sun, Will Hathaway, Neel Nanda, and Dimitris Bertsimas. 2024.
\newblock Universal neurons in gpt2 language models.
\newblock \emph{arXiv preprint arXiv:2401.12181}.

\bibitem[{Guu et~al.(2020)Guu, Lee, Tung, Pasupat, and Chang}]{guu2020retrieval}
Kelvin Guu, Kenton Lee, Zora Tung, Panupong Pasupat, and Mingwei Chang. 2020.
\newblock Retrieval augmented language model pre-training.
\newblock In \emph{International conference on machine learning}, pages 3929--3938. PMLR.

\bibitem[{Hawrylycz et~al.(2012)Hawrylycz, Lein, Guillozet-Bongaarts, Shen, Ng, Miller, Van De~Lagemaat, Smith, Ebbert, Riley et~al.}]{hawrylycz2012anatomically}
Michael~J Hawrylycz, Ed~S Lein, Angela~L Guillozet-Bongaarts, Elaine~H Shen, Lydia Ng, Jeremy~A Miller, Louie~N Van De~Lagemaat, Kimberly~A Smith, Amanda Ebbert, Zackery~L Riley, et~al. 2012.
\newblock An anatomically comprehensive atlas of the adult human brain transcriptome.
\newblock \emph{Nature}, 489(7416):391--399.

\bibitem[{Hendrycks et~al.()Hendrycks, Burns, Basart, Zou, Mazeika, Song, and Steinhardt}]{hendrycksmeasuring}
Dan Hendrycks, Collin Burns, Steven Basart, Andy Zou, Mantas Mazeika, Dawn Song, and Jacob Steinhardt.
\newblock Measuring massive multitask language understanding.
\newblock In \emph{International Conference on Learning Representations}.

\bibitem[{Hu et~al.()Hu, Wallis, Allen-Zhu, Li, Wang, Wang, Chen et~al.}]{hulora}
Edward~J Hu, Phillip Wallis, Zeyuan Allen-Zhu, Yuanzhi Li, Shean Wang, Lu~Wang, Weizhu Chen, et~al.
\newblock Lora: Low-rank adaptation of large language models.
\newblock In \emph{International Conference on Learning Representations}.

\bibitem[{Jiang et~al.(2024{\natexlab{a}})Jiang, Fang, Qiu, Zhang, Xu, Chen, Zhang, Zhang, Fang, Chu et~al.}]{jiang2024tc}
Xinke Jiang, Yue Fang, Rihong Qiu, Haoyu Zhang, Yongxin Xu, Hao Chen, Wentao Zhang, Ruizhe Zhang, Yuchen Fang, Xu~Chu, et~al. 2024{\natexlab{a}}.
\newblock Tc-rag: Turing-complete rag's case study on medical llm systems.
\newblock \emph{arXiv preprint arXiv:2408.09199}.

\bibitem[{Jiang et~al.(2024{\natexlab{b}})Jiang, Zhang, Xu, Qiu, Fang, Wang, Tang, Ding, Chu, Zhao et~al.}]{jiang2024hykge}
Xinke Jiang, Ruizhe Zhang, Yongxin Xu, Rihong Qiu, Yue Fang, Zhiyuan Wang, Jinyi Tang, Hongxin Ding, Xu~Chu, Junfeng Zhao, et~al. 2024{\natexlab{b}}.
\newblock Hykge: A hypothesis knowledge graph enhanced framework for accurate and reliable medical llms responses.
\newblock \emph{arXiv preprint arXiv:2312.15883}.

\bibitem[{Jiang et~al.(2023)Jiang, Xu, Gao, Sun, Liu, Dwivedi-Yu, Yang, Callan, and Neubig}]{jiang2023active}
Zhengbao Jiang, Frank~F Xu, Luyu Gao, Zhiqing Sun, Qian Liu, Jane Dwivedi-Yu, Yiming Yang, Jamie Callan, and Graham Neubig. 2023.
\newblock Active retrieval augmented generation.
\newblock In \emph{Proceedings of the 2023 Conference on Empirical Methods in Natural Language Processing}, pages 7969--7992.

\bibitem[{Jin et~al.(2024{\natexlab{a}})Jin, Cao, Chen, Liu, Jiang, Xu, Qiuxia, and Zhao}]{jin2024tug}
Zhuoran Jin, Pengfei Cao, Yubo Chen, Kang Liu, Xiaojian Jiang, Jiexin Xu, Li~Qiuxia, and Jun Zhao. 2024{\natexlab{a}}.
\newblock Tug-of-war between knowledge: Exploring and resolving knowledge conflicts in retrieval-augmented language models.
\newblock In \emph{Proceedings of the 2024 Joint International Conference on Computational Linguistics, Language Resources and Evaluation (LREC-COLING 2024)}, pages 16867--16878.

\bibitem[{Jin et~al.(2024{\natexlab{b}})Jin, Cao, Yuan, Chen, Xu, Li, Jiang, Liu, and Zhao}]{jin2024cutting}
Zhuoran Jin, Pengfei Cao, Hongbang Yuan, Yubo Chen, Jiexin Xu, Huaijun Li, Xiaojian Jiang, Kang Liu, and Jun Zhao. 2024{\natexlab{b}}.
\newblock Cutting off the head ends the conflict: A mechanism for interpreting and mitigating knowledge conflicts in language models.
\newblock In \emph{Findings of the Association for Computational Linguistics ACL 2024}, pages 1193--1215.

\bibitem[{Khandelwal et~al.(2019)Khandelwal, Levy, Jurafsky, Zettlemoyer, and Lewis}]{khandelwal2019generalization}
Urvashi Khandelwal, Omer Levy, Dan Jurafsky, Luke Zettlemoyer, and Mike Lewis. 2019.
\newblock Generalization through memorization: Nearest neighbor language models.
\newblock \emph{arXiv preprint arXiv:1911.00172}.

\bibitem[{Lewis et~al.(2020)Lewis, Perez, Piktus, Petroni, Karpukhin, Goyal, K{\"u}ttler, Lewis, Yih, Rockt{\"a}schel et~al.}]{lewis2020retrieval}
Patrick Lewis, Ethan Perez, Aleksandra Piktus, Fabio Petroni, Vladimir Karpukhin, Naman Goyal, Heinrich K{\"u}ttler, Mike Lewis, Wen-tau Yih, Tim Rockt{\"a}schel, et~al. 2020.
\newblock Retrieval-augmented generation for knowledge-intensive nlp tasks.
\newblock \emph{Advances in Neural Information Processing Systems}, 33:9459--9474.

\bibitem[{Li et~al.(2023{\natexlab{a}})Li, Rawat, Zaheer, Wang, Lukasik, Veit, Yu, and Kumar}]{li2023large}
Daliang Li, Ankit~Singh Rawat, Manzil Zaheer, Xin Wang, Michal Lukasik, Andreas Veit, Felix Yu, and Sanjiv Kumar. 2023{\natexlab{a}}.
\newblock Large language models with controllable working memory.
\newblock In \emph{Findings of the Association for Computational Linguistics: ACL 2023}, pages 1774--1793.

\bibitem[{Li et~al.(2023{\natexlab{b}})Li, Wang, Wu, Zhang, Xu, Fu, Tiwari, Wan, and Wang}]{li2023huatuo}
Jianquan Li, Xidong Wang, Xiangbo Wu, Zhiyi Zhang, Xiaolong Xu, Jie Fu, Prayag Tiwari, Xiang Wan, and Benyou Wang. 2023{\natexlab{b}}.
\newblock Huatuo-26m, a large-scale chinese medical qa dataset.
\newblock \emph{arXiv preprint arXiv:2305.01526}.

\bibitem[{Lin et~al.(2022)Lin, Hilton, and Evans}]{lin2022truthfulqa}
Stephanie Lin, Jacob Hilton, and Owain Evans. 2022.
\newblock Truthfulqa: Measuring how models mimic human falsehoods.
\newblock In \emph{Proceedings of the 60th Annual Meeting of the Association for Computational Linguistics (Volume 1: Long Papers)}, pages 3214--3252.

\bibitem[{Lin et~al.(2024)Lin, Ma, Chu, Jin, Yang, Wang, and Mei}]{lin2024lora}
Yang Lin, Xinyu Ma, Xu~Chu, Yujie Jin, Zhibang Yang, Yasha Wang, and Hong Mei. 2024.
\newblock Lora dropout as a sparsity regularizer for overfitting control.
\newblock \emph{arXiv preprint arXiv:2404.09610}.

\bibitem[{Liu et~al.(2024)Liu, Yao, Lv, Fan, Cao, Yu, Hou, and Li}]{liu2024untangle}
Yantao Liu, Zijun Yao, Xin Lv, Yuchen Fan, Shulin Cao, Jifan Yu, Lei Hou, and Juanzi Li. 2024.
\newblock Untangle the knot: Interweaving conflicting knowledge and reasoning skills in large language models.
\newblock In \emph{Proceedings of the 2024 Joint International Conference on Computational Linguistics, Language Resources and Evaluation (LREC-COLING 2024)}, pages 17186--17204.

\bibitem[{Ma et~al.(2024)Ma, Chu, Yang, Lin, Gao, and Zhao}]{pmlr-v235-ma24a}
Xinyu Ma, Xu~Chu, Zhibang Yang, Yang Lin, Xin Gao, and Junfeng Zhao. 2024.
\newblock Parameter efficient quasi-orthogonal fine-tuning via givens rotation.
\newblock In \emph{Proceedings of the 41st International Conference on Machine Learning}, volume 235, pages 33686--33729. PMLR.

\bibitem[{Ma et~al.(2023)Ma, Wang, Chu, Ma, Tang, Zhao, Yuan, and Wang}]{9992086}
Xinyu Ma, Yasha Wang, Xu~Chu, Liantao Ma, Wen Tang, Junfeng Zhao, Ye~Yuan, and Guoren Wang. 2023.
\newblock \href {https://doi.org/10.1109/TKDE.2022.3230454} {Patient health representation learning via correlational sparse prior of medical features}.
\newblock \emph{IEEE Transactions on Knowledge and Data Engineering}, 35(11):11769--11783.

\bibitem[{Ma et~al.(2025)Ma, Xu, Lin, Wang, Chu, Gao, Zhao, and Wang}]{ma2025dressing}
Xinyu Ma, Yifeng Xu, Yang Lin, Tianlong Wang, Xu~Chu, Xin Gao, Junfeng Zhao, and Yasha Wang. 2025.
\newblock \href {https://openreview.net/forum?id=mNVR9jJYqK} {{DRESS}ing up {LLM}: Efficient stylized question-answering via style subspace editing}.
\newblock In \emph{The Thirteenth International Conference on Learning Representations}.

\bibitem[{Mallen et~al.(2023)Mallen, Asai, Zhong, Das, Khashabi, and Hajishirzi}]{mallen2023not}
Alex Mallen, Akari Asai, Victor Zhong, Rajarshi Das, Daniel Khashabi, and Hannaneh Hajishirzi. 2023.
\newblock When not to trust language models: Investigating effectiveness of parametric and non-parametric memories.
\newblock In \emph{Proceedings of the 61st Annual Meeting of the Association for Computational Linguistics (Volume 1: Long Papers)}, pages 9802--9822.

\bibitem[{Meng et~al.(2022)Meng, Bau, Andonian, and Belinkov}]{meng2022locating}
Kevin Meng, David Bau, Alex Andonian, and Yonatan Belinkov. 2022.
\newblock Locating and editing factual associations in gpt.
\newblock \emph{Advances in Neural Information Processing Systems}, 35:17359--17372.

\bibitem[{Molchanov et~al.(2019)Molchanov, Mallya, Tyree, Frosio, and Kautz}]{molchanov2019importance}
Pavlo Molchanov, Arun Mallya, Stephen Tyree, Iuri Frosio, and Jan Kautz. 2019.
\newblock Importance estimation for neural network pruning.
\newblock In \emph{Proceedings of the IEEE/CVF conference on computer vision and pattern recognition}, pages 11264--11272.

\bibitem[{Nair and Hinton(2010)}]{nair2010rectified}
Vinod Nair and Geoffrey~E Hinton. 2010.
\newblock Rectified linear units improve restricted boltzmann machines.
\newblock In \emph{Proceedings of the 27th international conference on machine learning (ICML-10)}, pages 807--814.

\bibitem[{Olton et~al.(1979)Olton, Becker, and Handelmann}]{olton1979hippocampus}
David~S Olton, James~T Becker, and Gail~E Handelmann. 1979.
\newblock Hippocampus, space, and memory.
\newblock \emph{Behavioral and Brain sciences}, 2(3):313--322.

\bibitem[{Peng et~al.(2023)Peng, Galley, He, Cheng, Xie, Hu, Huang, Liden, Yu, Chen et~al.}]{peng2023check}
Baolin Peng, Michel Galley, Pengcheng He, Hao Cheng, Yujia Xie, Yu~Hu, Qiuyuan Huang, Lars Liden, Zhou Yu, Weizhu Chen, et~al. 2023.
\newblock Check your facts and try again: Improving large language models with external knowledge and automated feedback.
\newblock \emph{arXiv preprint arXiv:2302.12813}.

\bibitem[{Puig and Gulledge(2011)}]{puig2011serotonin}
M~Victoria Puig and Allan~T Gulledge. 2011.
\newblock Serotonin and prefrontal cortex function: neurons, networks, and circuits.
\newblock \emph{Molecular neurobiology}, 44:449--464.

\bibitem[{Rajpurkar et~al.(2018)Rajpurkar, Jia, and Liang}]{rajpurkar2018knowdontknowunanswerable}
Pranav Rajpurkar, Robin Jia, and Percy Liang. 2018.
\newblock \href {https://arxiv.org/abs/1806.03822} {Know what you don't know: Unanswerable questions for squad}.
\newblock \emph{Preprint}, arXiv:1806.03822.

\bibitem[{Ren et~al.(2023)Ren, Wang, Qu, Zhao, Liu, Tian, Wu, Wen, and Wang}]{ren2023investigating}
Ruiyang Ren, Yuhao Wang, Yingqi Qu, Wayne~Xin Zhao, Jing Liu, Hao Tian, Hua Wu, Ji-Rong Wen, and Haifeng Wang. 2023.
\newblock Investigating the factual knowledge boundary of large language models with retrieval augmentation.
\newblock \emph{arXiv preprint arXiv:2307.11019}.

\bibitem[{Rizzolatti and Craighero(2004)}]{rizzolatti2004mirror}
Giacomo Rizzolatti and Laila Craighero. 2004.
\newblock The mirror-neuron system.
\newblock \emph{Annu. Rev. Neurosci.}, 27(1):169--192.

\bibitem[{Sakaguchi et~al.(2021)Sakaguchi, Bras, Bhagavatula, and Choi}]{sakaguchi2021winogrande}
Keisuke Sakaguchi, Ronan~Le Bras, Chandra Bhagavatula, and Yejin Choi. 2021.
\newblock Winogrande: An adversarial winograd schema challenge at scale.
\newblock \emph{Communications of the ACM}, 64(9):99--106.

\bibitem[{Shi et~al.(2024{\natexlab{a}})Shi, Jin, Shen, Dong, Wu, and Xiong}]{shi2024ircan}
Dan Shi, Renren Jin, Tianhao Shen, Weilong Dong, Xinwei Wu, and Deyi Xiong. 2024{\natexlab{a}}.
\newblock Ircan: Mitigating knowledge conflicts in llm generation via identifying and reweighting context-aware neurons.
\newblock \emph{arXiv preprint arXiv:2406.18406}.

\bibitem[{Shi et~al.(2023)Shi, Chen, Misra, Scales, Dohan, Chi, Sch{\"a}rli, and Zhou}]{shi2023large}
Freda Shi, Xinyun Chen, Kanishka Misra, Nathan Scales, David Dohan, Ed~H Chi, Nathanael Sch{\"a}rli, and Denny Zhou. 2023.
\newblock Large language models can be easily distracted by irrelevant context.
\newblock In \emph{International Conference on Machine Learning}, pages 31210--31227. PMLR.

\bibitem[{Shi et~al.(2024{\natexlab{b}})Shi, Han, Lewis, Tsvetkov, Zettlemoyer, and Yih}]{shi2024trusting}
Weijia Shi, Xiaochuang Han, Mike Lewis, Yulia Tsvetkov, Luke Zettlemoyer, and Wen-tau Yih. 2024{\natexlab{b}}.
\newblock Trusting your evidence: Hallucinate less with context-aware decoding.
\newblock In \emph{Proceedings of the 2024 Conference of the North American Chapter of the Association for Computational Linguistics: Human Language Technologies (Volume 2: Short Papers)}, pages 783--791.

\bibitem[{Su et~al.(2024)Su, Tang, Ai, Wu, and Liu}]{su2024dragin}
Weihang Su, Yichen Tang, Qingyao Ai, Zhijing Wu, and Yiqun Liu. 2024.
\newblock Dragin: Dynamic retrieval augmented generation based on the real-time information needs of large language models.
\newblock \emph{arXiv preprint arXiv:2403.10081}.

\bibitem[{Tan et~al.(2024)Tan, Sun, Yang, Wang, Cao, and Cheng}]{tan2024blinded}
Hexiang Tan, Fei Sun, Wanli Yang, Yuanzhuo Wang, Qi~Cao, and Xueqi Cheng. 2024.
\newblock Blinded by generated contexts: How language models merge generated and retrieved contexts for open-domain qa?
\newblock \emph{arXiv preprint arXiv:2401.11911}.

\bibitem[{Tan et~al.(2023)Tan, Ng, and Bing}]{tan2023towards}
Qingyu Tan, Hwee~Tou Ng, and Lidong Bing. 2023.
\newblock Towards benchmarking and improving the temporal reasoning capability of large language models.
\newblock In \emph{Proceedings of the 61st Annual Meeting of the Association for Computational Linguistics (Volume 1: Long Papers)}, pages 14820--14835.

\bibitem[{Tang et~al.(2024)Tang, Luo, Huang, Zhang, Wang, Zhao, Wei, and Wen}]{tang2024language}
Tianyi Tang, Wenyang Luo, Haoyang Huang, Dongdong Zhang, Xiaolei Wang, Xin Zhao, Furu Wei, and Ji-Rong Wen. 2024.
\newblock Language-specific neurons: The key to multilingual capabilities in large language models.
\newblock \emph{arXiv preprint arXiv:2402.16438}.

\bibitem[{Team(2024)}]{team2024introducing}
Qwen Team. 2024.
\newblock Introducing qwen1. 5.

\bibitem[{Touvron et~al.(2023)Touvron, Martin, Stone, Albert, Almahairi, Babaei, Bashlykov, Batra, Bhargava, Bhosale et~al.}]{touvron2023llama}
Hugo Touvron, Louis Martin, Kevin Stone, Peter Albert, Amjad Almahairi, Yasmine Babaei, Nikolay Bashlykov, Soumya Batra, Prajjwal Bhargava, Shruti Bhosale, et~al. 2023.
\newblock Llama 2: Open foundation and fine-tuned chat models.
\newblock \emph{arXiv preprint arXiv:2307.09288}.

\bibitem[{Trivedi et~al.(2022)Trivedi, Balasubramanian, Khot, and Sabharwal}]{trivedi2021musique}
Harsh Trivedi, Niranjan Balasubramanian, Tushar Khot, and Ashish Sabharwal. 2022.
\newblock {M}u{S}i{Q}ue: Multihop questions via single-hop question composition.
\newblock \emph{Transactions of the Association for Computational Linguistics}.

\bibitem[{Wang et~al.(2024{\natexlab{a}})Wang, Yao, Xu, Qiao, Deng, Wang, Chen, Gu, Jiang, Xie et~al.}]{wang2024knowledge}
Mengru Wang, Yunzhi Yao, Ziwen Xu, Shuofei Qiao, Shumin Deng, Peng Wang, Xiang Chen, Jia-Chen Gu, Yong Jiang, Pengjun Xie, et~al. 2024{\natexlab{a}}.
\newblock Knowledge mechanisms in large language models: A survey and perspective.
\newblock \emph{arXiv preprint arXiv:2407.15017}.

\bibitem[{Wang et~al.(2023{\natexlab{a}})Wang, Zhu, Liu, Zheng, Chen et~al.}]{wang2023knowledge}
Song Wang, Yaochen Zhu, Haochen Liu, Zaiyi Zheng, Chen Chen, et~al. 2023{\natexlab{a}}.
\newblock Knowledge editing for large language models: A survey.
\newblock \emph{arXiv preprint arXiv:2310.16218}.

\bibitem[{Wang et~al.(2023{\natexlab{b}})Wang, Chen, Ge, Xia, Bao, Zheng, Zhang, Gui, and Huang}]{wang2023orthogonal}
Xiao Wang, Tianze Chen, Qiming Ge, Han Xia, Rong Bao, Rui Zheng, Qi~Zhang, Tao Gui, and Xuan-Jing Huang. 2023{\natexlab{b}}.
\newblock Orthogonal subspace learning for language model continual learning.
\newblock In \emph{Findings of the Association for Computational Linguistics: EMNLP 2023}, pages 10658--10671.

\bibitem[{Wang et~al.(2024{\natexlab{b}})Wang, Chen, Dingjie, Zhiyi, Chen, Xiao, Chen, Jiang, Li, Wan et~al.}]{wang2024cmb}
Xidong Wang, Guiming Chen, Song Dingjie, Zhang Zhiyi, Zhihong Chen, Qingying Xiao, Junying Chen, Feng Jiang, Jianquan Li, Xiang Wan, et~al. 2024{\natexlab{b}}.
\newblock Cmb: A comprehensive medical benchmark in chinese.
\newblock In \emph{Proceedings of the 2024 Conference of the North American Chapter of the Association for Computational Linguistics: Human Language Technologies (Volume 1: Long Papers)}, pages 6184--6205.

\bibitem[{Wolf(2019)}]{wolf2019huggingface}
T~Wolf. 2019.
\newblock Huggingface's transformers: State-of-the-art natural language processing.
\newblock \emph{arXiv preprint arXiv:1910.03771}.

\bibitem[{Wu et~al.(2024{\natexlab{a}})Wu, Wu, and Zou}]{wu2024clasheval}
Kevin Wu, Eric Wu, and James Zou. 2024{\natexlab{a}}.
\newblock Clasheval: Quantifying the tug-of-war between an llm’s internal prior and external evidence.
\newblock \emph{Preprint}.

\bibitem[{Wu et~al.(2024{\natexlab{b}})Wu, Wu, and Zou}]{wu2024faithful}
Kevin Wu, Eric Wu, and James Zou. 2024{\natexlab{b}}.
\newblock How faithful are rag models? quantifying the tug-of-war between rag and llms' internal prior.
\newblock \emph{arXiv preprint arXiv:2404.10198}.

\bibitem[{Xiao et~al.(2023)Xiao, Liu, Zhang, Muennighoff, Lian, and Nie}]{xiao2023c}
Shitao Xiao, Zheng Liu, Peitian Zhang, Niklas Muennighoff, Defu Lian, and Jian-Yun Nie. 2023.
\newblock C-pack: Packaged resources to advance general chinese embedding.
\newblock \emph{arXiv preprint arXiv:2309.07597}.

\bibitem[{Xie et~al.()Xie, Zhang, Chen, Lou, and Su}]{xieadaptive}
Jian Xie, Kai Zhang, Jiangjie Chen, Renze Lou, and Yu~Su.
\newblock Adaptive chameleon or stubborn sloth: Revealing the behavior of large language models in knowledge conflicts.
\newblock In \emph{The Twelfth International Conference on Learning Representations}.

\bibitem[{Xu et~al.(2024{\natexlab{a}})Xu, Pang, Yu, Meng, Shen, Cheng, and Zhou}]{xu2024unsupervised}
Shicheng Xu, Liang Pang, Mo~Yu, Fandong Meng, Huawei Shen, Xueqi Cheng, and Jie Zhou. 2024{\natexlab{a}}.
\newblock Unsupervised information refinement training of large language models for retrieval-augmented generation.
\newblock \emph{arXiv preprint arXiv:2402.18150}.

\bibitem[{Xu et~al.(2023{\natexlab{a}})Xu, Chu, Yang, Wang, Zou, Ding, Zhao, Wang, and Xie}]{xu2023seqcare}
Yongxin Xu, Xu~Chu, Kai Yang, Zhiyuan Wang, Peinie Zou, Hongxin Ding, Junfeng Zhao, Yasha Wang, and Bing Xie. 2023{\natexlab{a}}.
\newblock Seqcare: Sequential training with external medical knowledge graph for diagnosis prediction in healthcare data.
\newblock In \emph{Proceedings of the ACM Web Conference 2023}, pages 2819--2830.

\bibitem[{Xu et~al.(2025)Xu, Jiang, Chu, Qiu, Feng, Ding, Zhao, Wang, and Xie}]{xu2025dearllm}
Yongxin Xu, Xinke Jiang, Xu~Chu, Rihong Qiu, Yujie Feng, Hongxin Ding, Junfeng Zhao, Yasha Wang, and Bing Xie. 2025.
\newblock Dearllm: Enhancing personalized healthcare via large language models-deduced feature correlations.
\newblock In \emph{Proceedings of the AAAI Conference on Artificial Intelligence}, volume~39, pages 941--949.

\bibitem[{Xu et~al.(2024{\natexlab{b}})Xu, Jiang, Chu, Xiao, Zhang, Ding, Zhao, Wang, and Xie}]{xu2024protomix}
Yongxin Xu, Xinke Jiang, Xu~Chu, Yuzhen Xiao, Chaohe Zhang, Hongxin Ding, Junfeng Zhao, Yasha Wang, and Bing Xie. 2024{\natexlab{b}}.
\newblock Protomix: Augmenting health status representation learning via prototype-based mixup.
\newblock In \emph{Proceedings of the 30th ACM SIGKDD Conference on Knowledge Discovery and Data Mining}, pages 3633--3644.

\bibitem[{Xu et~al.(2023{\natexlab{b}})Xu, Yang, Zhang, Zou, Wang, Ding, Zhao, Wang, and Xie}]{xu2023vecocare}
Yongxin Xu, Kai Yang, Chaohe Zhang, Peinie Zou, Zhiyuan Wang, Hongxin Ding, Junfeng Zhao, Yasha Wang, and Bing Xie. 2023{\natexlab{b}}.
\newblock Vecocare: Visit sequences-clinical notes joint learning for diagnosis prediction in healthcare data.
\newblock In \emph{IJCAI}, volume~23, pages 4921--4929.

\bibitem[{Yang et~al.(2023)Yang, Xu, Zou, Ding, Zhao, Wang, and Xie}]{yang2023kerprint}
Kai Yang, Yongxin Xu, Peinie Zou, Hongxin Ding, Junfeng Zhao, Yasha Wang, and Bing Xie. 2023.
\newblock Kerprint: local-global knowledge graph enhanced diagnosis prediction for retrospective and prospective interpretations.
\newblock In \emph{Proceedings of the AAAI Conference on Artificial Intelligence}, volume~37, pages 5357--5365.

\bibitem[{Yoran et~al.()Yoran, Wolfson, Ram, and Berant}]{yoranmaking}
Ori Yoran, Tomer Wolfson, Ori Ram, and Jonathan Berant.
\newblock Making retrieval-augmented language models robust to irrelevant context.
\newblock In \emph{The Twelfth International Conference on Learning Representations}.

\bibitem[{Zellers et~al.(2019)Zellers, Holtzman, Bisk, Farhadi, and Choi}]{zellers2019hellaswag}
Rowan Zellers, Ari Holtzman, Yonatan Bisk, Ali Farhadi, and Yejin Choi. 2019.
\newblock Hellaswag: Can a machine really finish your sentence?
\newblock In \emph{Proceedings of the 57th Annual Meeting of the Association for Computational Linguistics}, pages 4791--4800.

\bibitem[{Zhang et~al.(2023)Zhang, Chen, Bukharin, Karampatziakis, He, Cheng, Chen, and Zhao}]{zhang2023adalora}
Qingru Zhang, Minshuo Chen, Alexander Bukharin, Nikos Karampatziakis, Pengcheng He, Yu~Cheng, Weizhu Chen, and Tuo Zhao. 2023.
\newblock Adalora: Adaptive budget allocation for parameter-efficient fine-tuning.
\newblock \emph{arXiv preprint arXiv:2303.10512}.

\bibitem[{Zhang et~al.(2022)Zhang, Zuo, Liang, Bukharin, He, Chen, and Zhao}]{zhang2022platon}
Qingru Zhang, Simiao Zuo, Chen Liang, Alexander Bukharin, Pengcheng He, Weizhu Chen, and Tuo Zhao. 2022.
\newblock Platon: Pruning large transformer models with upper confidence bound of weight importance.
\newblock In \emph{International conference on machine learning}, pages 26809--26823. PMLR.

\bibitem[{Zhang et~al.(2025)Zhang, Xu, Xiao, Zhu, Jiang, Chu, Zhao, and Wang}]{zhang2025knowpo}
Ruizhe Zhang, Yongxin Xu, Yuzhen Xiao, Runchuan Zhu, Xinke Jiang, Xu~Chu, Junfeng Zhao, and Yasha Wang. 2025.
\newblock Knowpo: Knowledge-aware preference optimization for controllable knowledge selection in retrieval-augmented language models.
\newblock In \emph{Proceedings of the AAAI Conference on Artificial Intelligence}, volume~39, pages 25895--25903.

\bibitem[{Zhao et~al.(2023)Zhao, Li, Joty, Qin, and Bing}]{zhao2023verify}
Ruochen Zhao, Xingxuan Li, Shafiq Joty, Chengwei Qin, and Lidong Bing. 2023.
\newblock Verify-and-edit: A knowledge-enhanced chain-of-thought framework.
\newblock In \emph{Proceedings of the 61st Annual Meeting of the Association for Computational Linguistics (Volume 1: Long Papers)}, pages 5823--5840.

\bibitem[{Zhou et~al.(2023)Zhou, Zhang, Poon, and Chen}]{zhou2023context}
Wenxuan Zhou, Sheng Zhang, Hoifung Poon, and Muhao Chen. 2023.
\newblock Context-faithful prompting for large language models.
\newblock In \emph{Findings of the Association for Computational Linguistics: EMNLP 2023}, pages 14544--14556.

\end{thebibliography}

\appendix
\section{Appendix}
\label{sec:appendix}
\begin{algorithm}[htbp]
\footnotesize
\caption{Training Algorithm of \mname}\label{alg:training}
\begin{algorithmic}[1]
\renewcommand{\algorithmicrequire}{\textbf{Input:}}
\renewcommand{\algorithmicensure}{\textbf{Output:}}
\REQUIRE Training dataset $S_a$, $S_r$, and $S_c$; total probing iterations $T_{prob}$; total training iterations $T_{tr}$; initial LLMs $\Theta$.

\FOR {$* \in \left\{a,r\right\}$}
\FOR {$B$ in mini-batches of $\mathcal{C}_*$}
    \STATE Calculate and store the activation value.
\ENDFOR
\STATE Calculate expected activation value $p_{*,k}^j$ via Eq. (\ref{eq:single_activation}).
\STATE Obtain the sensitivity \( \hat{p}_*^j \) through aggregation, for $j \in\{1, \ldots, l\}$ via Eq. (\ref{eq:average_activation}).
\ENDFOR

\FOR {$* \in \left\{a,r\right\}$}
\FOR {$t = 1, \ldots, T_{prob}$}
    \STATE Sample a mini-batch from $\mathcal{S}_*$  and compute the gradient $\nabla \mathcal{L}_*$;
    \STATE Compute the sensitivity $I_*(w_{uv})$ for each parameter via Eq. (\ref{eq:parameter_importance});
    \STATE Update the importance score $\overline{s}_*^{(t)}(w_{uv})$ for each parameter via Eq. (\ref{eq:ema_uncertainty}) and (\ref{eq:final_importance});
\ENDFOR
\STATE Obtain the importance distribution $\mathcal{I}_*(\mathcal{E})$ through aggregation.
\ENDFOR

\STATE Obtain $\mathcal{E}_c$, $\mathcal{E}_{ax}$, and $\mathcal{E}_{rx}$ respectively using Eq. (\ref{eq:z_score_intersecting}), (\ref{eq:z_score_adherence}), and (\ref{eq:z_score_robustness}).

\FOR {$t = 1, \ldots, T_{tr}$}
    \STATE Update $\mathcal{E}_c$ by optimizing $\mathcal{L}_c$ based on Eq. (\ref{eq: loss-c});
    \STATE Update $\mathcal{E}_{ax}$ by optimizing $\mathcal{L}_{ax}$ based on Eq. (\ref{eq: loss-ax});
    \STATE Update $\mathcal{E}_{rx}$ by optimizing $\mathcal{L}_{rx}$ based on Eq. (\ref{eq: loss-rx});
\ENDFOR
\ENSURE The fine-tuned LLMs $\Theta^\prime$.

\end{algorithmic} 
\end{algorithm}

\subsection{\mname Algorithm Framework}
\label{appendix:algorithm}
The detailed implementation of \mname algorithm can be found in Algorithm~\ref{alg:training}.

\subsection{Notations}
\label{sec:app notation}
The notations in this paper are summarized in Table~\ref{tab:symbols}.

\begin{table}[htb]
    \footnotesize
    \centering
    \begin{tabular}{c|l}        
        \hline
        \rowcolor[gray]{0.94} Notation & Definition \\ 
        \hline
        $\Theta$ & Initial LLMs \\
        $\Theta^\prime$ & LLMs fine-tuned using \mname\\
        $e_i$ & A single parameter unit in the model \\
        $\mathcal{E}$ & The set of parameter units in the model, \\
        & $\mathcal{E} = \{e_1, e_2, \dots, e_n\}$ \\
        $\alpha$ & Internal parametric knowledge of $\Theta$ \\ 
        
        \hdashline
        $q$ & Input natural language query \\ 
        $D$ & The set of retrieved documents for $q$, \\
        & $D = \{d_1, d_2, \dots, d_k\}$ \\ 
        $d_{\text{golden}}$ & Evidence required to answer query $q$ \\ 
        $y$ & Relevance label indicating whether \\
        & the retrieved documents contain evidence \\
        $\mathbb{C}$ & External knowledge base \\
        $ans_{golden}$ & Correct answer derived from $d_{golden}$ \\
        \hdashline
        $S_a$ & SFT dataset promoting adherence \\ 
        $S_r$ & SFT dataset promoting robustness \\
        $S_c $ & Document extraction dataset \\
        $\mathcal{L}_*$,$* \in \left\{a,r,c\right\}$ & Cross-entropy loss on $S_*$ \\
        $\mathcal{L}_{ax}$ & Loss for the adherence subspace\\ 
        $\mathcal{L}_{rx}$ & Loss for the robustness subspace\\ 
        $\mathcal{L}_{cx}$ & Loss for the entangled subspace \\  
        \hdashline
        $p_*^j$ & Expected activation probability of layer $j$\\
        $\overline{I}_* / \overline{U}_*(w_{uv})$, & Sensitivity / Uncertainty score for \\
        $* \in \left\{a,r\right\}$ & parameter $w_{uv}$ based on gradients \\ 
        $\overline{s}_*(w_{uv})$ & Final importance score for parameter $w_{uv}$ \\ 
        $\mathcal{I}_*(\mathcal{E})$ & Distribution of importance scores\\
        $\mathcal{Z}_*(e_i)$ & Standardized importance score of the \\
        & \( i \)-th parameter unit \\ 
        \hdashline 
        $\mathcal{E}_c$ & Entangled subspace \\ 
        $\mathcal{E}_{ax}$ & Adherence subspace \\ 
        $\mathcal{E}_{rx}$ & Robustness subspace \\ 
        $\mathcal{E}_{o}$ & Other subspaces \\ 
        \hline
    \end{tabular}
    \caption{\label{tab:symbols}Notations for \mname}
\end{table}

\subsection{Dataset Details}
\label{appedix:data_details}
\begin{itemize}[leftmargin=*,noitemsep,topsep=3pt]
\item \textbf{SQuAD2.0}~\cite{rajpurkar2018knowdontknowunanswerable} is a reading comprehension dataset encompassing multiple general domains, consisting of questions posed by crowdworkers on a set of Wikipedia articles, where the answer to every question is a segment of text, or span, from the corresponding reading passage, or the question might be unanswerable.

Following the methodology outlined in Section~\ref{section:sft}, we construct $S_a$ and $S_r$ using the SQuAD2.0 dataset, with each containing 6,000 entries, and each entry consisting of a single document. 
For $S_a$, the document serves as either evidence or as a resource that could help produce a fabricated answer conflicting with the model's internal knowledge. 
During the process of expanding the dataset, we use GPT-4 to generate fictional answers that deviate from the true answers. We then replace all occurrences of the true answers in the evidence documents with these fictional ones to effectively augment the dataset. 
This is because the essence of knowledge conflict lies in the conflict itself, rather than in correctness~\cite{li2023large}.
For $S_r$, the document consists of manually annotated irrelevant noise.
Following the methodology described in Section~\ref{section:evidence}, we develop $S_c$, which comprises 12,000 data entries.  Each entry includes four documents: one document relevant to the question's answer (both correct and fabricated), one noise document on the same topic, and two noise documents from entirely different topics.

Following~\cite{li2023large}, we create the evaluation set SQuAD2.0-Eval to simulate complex retrieval scenarios. We construct two types of contexts: conflicting contexts and irrelevant contexts. In line with the typical chunk-size used in RAG tasks~\cite{shi2023large}, we set the number of documents in each context to four. 
For the conflicting context, we select one document related to the same topic and two documents from different topics, all based on semantic similarity (We utilize the widely adopted and advanced bge-reranker-large~\cite{xiao2023c} model.). We ensure that none of these documents can provide an answer to the question. These are then paired with an evidence document that directly conflicts with the internal knowledge of the LLMs.
To prevent bias arising from the order of documents, we randomize the order of evidence within the context.
For the irrelevant context, we differentiate the documents by difficulty level. The more challenging documents, manually annotated, include two topic-related but unanswerable documents; the less challenging are randomly selected, consisting of two topic-unrelated documents. These four documents are then shuffled to form the irrelevant context.

\item \textbf{RGB}~\cite{wang2024cmb} serves as a benchmark to evaluate LLMs' ability to effectively utilize retrieved information and withstand various retrieval-related flaws, covering both English and Chinese languages. RGB provides question-answer pairs annotated with counterfactual and noisy knowledge. Consistent with our approach in constructing SQuAD2.0-Eval, we build an evaluation dataset with more complex contexts based on the original corpus.

\item \textbf{KNOT}~\cite{liu2024untangle} is a benchmark specifically designed for studying the resolution of knowledge conflicts. It categorizes questions into three types based on the reasoning capabilities required to reconcile conflicting information. Consistent with our methodology in constructing SQuAD2.0-Eval, we further enrich this foundation by incorporating noisy contexts, and creating an evaluation dataset with more complex scenarios to test performance.

\item \textbf{CMB}~\cite{wang2024cmb} is a medical open-source query dataset, which are designed for multi-task Q\&A, encompass single and multiple-choice questions in the medical field~\cite{jiang2024hykge}. 
The CMB dataset utilizes qualifying exams as a data source in the four clinical medicine specialties of physicians, nurses, medical technicians, and pharmacists, with a total of 269,359 questions. Given the extensive size of the CMB dataset, we randomly sample 4,000 questions for testing.
The medical domain concerns human health and life safety~\cite{yang2023kerprint,xu2023seqcare,9992086,xu2023vecocare,xu2024protomix}. Developing language models capable of understanding and applying medical knowledge is of great significance for advancing the intelligence of healthcare services.
Consistent with our approach for constructing the SQuAD2.0-Eval, by utilizing medical triples and documents from the Huatuo-26M dataset~\cite{li2023huatuo} to build supplementary medical context, we create a dataset to evaluate the medical knowledge utilization of LLMs.
Specifically, we construct conflicting knowledge and contexts with original questions and answers, as well as irrelevant contexts. We do not require LLMs to complete multiple-choice questions but to directly answer the questions.

\end{itemize}

\subsection{Prompts used in \mname}
\label{appendix:prompts}
In this section, we provide a detailed display of all the prompts used.
\begin{itemize}[leftmargin=*,noitemsep,topsep=3pt]
\item \textbf{Extract parameter knowledge}
The following prompt extracts world knowledge acquired during the pretraining phase of LLMs.
\begin{tcolorbox}[colback=lightgray!20,colframe=darkgray!80,title=Prompt A.4.1]
This is a question about \{\textit{Topic}\}. Please answer the question \{\textit{Question $q$}\}. Please provide a direct answer without analysis. If you are unsure or do not know the answer, please respond with `I don't know'.
\end{tcolorbox}

\item \textbf{Construct fabricated (counterfactual) answers}
The following few-shot prompt is used to generate fabricated answers using GPT-4
that deviates from the realistic answer, which we require to be as plausible as possible.
\begin{tcolorbox}[colback=lightgray!20,colframe=darkgray!80,title=Prompt A.4.2]
Please generate speciously plausible but incorrect answer to the question. Provide only the false answers; do not reiterate the queries.

Question: What is the capital of France? Answer: Paris. Fake answer: Lyon.

Question: What is the highest mountain in the world? Answer: Mount Everest. Fake answer: Lhotse.

\textit{\ldots 7 more examples \ldots}

Question: Who is the founder of Microsoft? Answer: Bill Gates. Fake answer: Steve Jobs.

Question: \{\textit{Question $q$}\} Answer: \{\textit{Realistic Answer $ans_{golden}$}\} Fake answer:
\end{tcolorbox}

\item \textbf{Guide LLMs to integrate knowledge}
The following prompt guides LLMs to integrate both external retrieved knowledge and internal parameter-based knowledge, serving as the input framework for the training sets $S_a$ and $S_r$, as well as for all evaluation sets.

\begin{tcolorbox}[colback=lightgray!20,colframe=darkgray!80,title=Prompt A.4.3]
\textbf{[Instruction]} \quad As a knowledge-based QA expert, you will provide professional responses based on user's question, utilizing any supplemental knowledge provided to enhance the quality of your response. If the supplemental information is irrelevant to the question, rely on your own expertise to formulate an answer. If you are unsure about the answer, please respond with `I don't know'.

\textbf{[Supplemental Knowledge]} \quad \{\textit{Context $D$}\}

\textbf{[Question]} \quad \{\textit{Question $q$}\}

\textbf{[Answer]}
\end{tcolorbox}

\item \textbf{Document Extraction}
The following prompt guides LLMs to identify specific document types and accurately restate their content.
\begin{tcolorbox}[colback=lightgray!20,colframe=darkgray!80,title=Prompt A.4.4]
\textbf{[Instruction]} \quad Based on the topic, question, and supplementary knowledge provided below, assess the relevance of the supplementary knowledge to the question and classify it. 

Output the document that is relevant to the topic and supports answering the question. \textit{or} [\textcolor{gray}{Output the document that is relevant to the topic but does not directly assist in answering the question.}] \textit{or} [\textcolor{gray}{Output the document that is unrelated to both the topic and the question.}]

\textbf{[Topic]} \quad \{\textit{Topic}\}

\textbf{[Supplemental Knowledge]} \quad \{\textit{Context $D$}\}

\textbf{[Question]} \quad \{\textit{Question $q$}\}

\textbf{[Answer]}
\end{tcolorbox}

\item \textbf{Noise recognition}
The following prompt directs LLMs to classify document types, serving as the input framework for the noise identification experiment described in Section~\ref{appendix:noise_rec}.
\begin{tcolorbox}[colback=lightgray!20,colframe=darkgray!80,title=Prompt A.4.5]
\textbf{[Instruction]} \quad Based on the question and document provided below, assess whether the document assists in answering the question. If the document contains the answer to the question, output the evidence; otherwise, output noise.

\textbf{[Supplemental Document]} \quad \{\textit{Document $d_k$}\}

\textbf{[Question]} \quad \{\textit{Question $q$}\}

\textbf{[Evaluation Result]}
\end{tcolorbox}

\end{itemize}

\subsection{Implementation Details}
\label{appendix:imple}
We employ two different parameter-level language model architectures: LLaMA2-7B-Chat~\cite{touvron2023llama} and Qwen1.5-14B-Chat~\cite{team2024introducing}. For both models, we utilize parameter-efficient fine-tuning techniques, specifically LoRA, to expedite the training process.
We use PyTorch library to implement all the algorithms based on the open-source HuggingFace transformers~\cite{wolf2019huggingface} codebase, on an Ubuntu server equipped with 4 NVIDIA A100 GPUs with 80GB memory. 
For \mname, we set the hyperparameters $\alpha_1$ and $\alpha_2$ in Equation (\ref{eq:ema_uncertainty}) to 0.85, and set \( \delta_1 \) in Equations (\ref{eq: loss-c}), (\ref{eq: loss-ax}), and (\ref{eq: loss-rx}) to 0.5.

During the probing phase (i.e., the key parameter mining stage described in Section~\ref{sec: Importance-aware parameter unit type recognition}), we use a learning rate of 1e-4, a batch size of 16, a cutoff length of 1024, and run the process for 1 epoch.

In the training phase, for LLaMA2-7B-Chat:  Training was conducted with a learning rate of 1e-4, a batch size of 16, a cutoff length of 1024, and 3 epochs. Additionally, LoRA adjustments were implemented with settings including a rank of 8, alpha of 16, and a dropout rate of 0.05, specifically targeting the modules [o\_proj,q\_proj,k\_proj,v\_proj]. 
For Qwen1.5-14B-Chat: Training was conducted with a learning rate of 5e-5, a batch size of 8, a cutoff length of 1024, and 3 epochs. LoRA settings were rank = 8, alpha = 16, dropout = 0.05, targeting modules [o\_proj,q\_proj,k\_proj,v\_proj]. 

For testing, settings included temperature = 0.02, top\_p = 0, top\_k = 1, max new tokens = 512. 

Additionally, we carefully tuned the hyper-parameters of baselines according to the suggestions in the original paper, in order to achieve their optimal performance.

\begin{table*}[t]
  \centering
  \resizebox{\textwidth}{!}{
    \begin{tabular}{lcccccccc}
    \toprule
    \multicolumn{2}{c}{Models} & ARC   & HellaSwag & MMLU  & TruthfulQA & Winogrande & GSM8K & Average \\
    \midrule
    \multirow{2}[2]{*}{LLaMA2-7B-Chat} & Original & 51.84  & 77.55  & 47.99  & 46.32  & 71.63  & 22.16  & 52.92  \\
          & \mname & 51.95  & 77.57  & 45.96  & 46.35  & 71.66  & 21.33  & 52.47  \\
    \midrule
    \multirow{2}[2]{*}{Qwen1.5-14B-Chat} & Original & 60.39  & 82.43  & 64.49  & 53.14  & 78.38  & 68.27  & 67.85  \\
          & \mname & 60.42  & 82.03  & 64.58  & 53.10  & 78.44  & 68.12  &  67.78 \\
    \bottomrule
    \end{tabular}
    }
    \caption{Results of general abilities of LLMs on six widely-used benchmarks.}
  \label{table:general_abilities}
\end{table*}

\begin{table}[t]
\centering
\begin{tabular}{lcccc}
\toprule
\multirow{2}{*}{$\alpha_1$,$\alpha_2$} & \multicolumn{2}{c}{SQuAD} & \multicolumn{2}{c}{RGB} \\
\cmidrule(lr){2-5}
                     & $R_{Ad}$          & $R_{Ro}$         & $R_{Ad}$         & $R_{Ro}$         \\
\midrule
0.15      &       65.57      &      40.12       &      78.00      &      42.50      \\
0.35     &      66.49       &       42.31      &      79.50      &      43.50      \\
0.55   &        67.62     &       43.28      &      78.00      &      44.50      \\
0.85   &      69.24       &     44.85  &   79.50      &   45.50     \\
0.95  &       67.01      &       42.51      &      78.50      &  43.00 \\       
\bottomrule
\end{tabular}
\caption{\label{table:senstivity_alpha}Performance comparisons of \mname (using LLaMA2-7B-Chat as the backbone model) across SQuAD2.0-Eval and RGB with different \( \alpha \) values configured.}
\end{table}

\begin{table}[t]
\centering
\begin{tabular}{lcccc}
\toprule
\multirow{2}{*}{$\delta_1$} & \multicolumn{2}{c}{SQuAD} & \multicolumn{2}{c}{RGB} \\
\cmidrule(lr){2-5}
                     & $R_{Ad}$          & $R_{Ro}$         & $R_{Ad}$         & $R_{Ro}$         \\
\midrule
0.1      &       61.56      &      36.93       &       69.50     &      36.50      \\
0.3     &      68.12       &      44.71       &      80.00      & 44.00            \\
0.5   &     69.24       &     44.85  &   79.50      &   45.50     \\
0.7   &      68.94       &   43.14    &    79.50    &    44.50   \\
0.9  &      63.57       &       37.54      &      75.50      &  37.50 \\       
\bottomrule
\end{tabular}
\caption{\label{table:senstivity_delta}Performance comparisons of \mname (using LLaMA2-7B-Chat as the backbone model) across SQuAD2.0-Eval and RGB with different $\delta_1$ values configured.}
\end{table}

\subsection{General Task Performance}
\label{appendix:general_performance}
Targeted optimization for RAG scenarios may unintentionally have side effects on the general capabilities of LLMs.
To comprehensively evaluate the impact of \mname, we conducted systematic assessments across six widely-used benchmarks, measuring its performance on various general tasks, including multitask language understanding, logical reasoning, and factuality. These benchmarks include ARC~\cite{clark2018think}, HellaSwag~\cite{zellers2019hellaswag}, MMLU~\cite{hendrycksmeasuring}, Winogrande~\cite{sakaguchi2021winogrande}, GSM8K~\cite{cobbe2021training}, and TruthfulQA~\cite{lin2022truthfulqa}.
Following the setup in~\citet{shi2024ircan}, we also used the Eleuther AI LM Evaluation Harness~$\footnote{https://github.com/EleutherAI/lm-evaluation-harness}$ for evaluation.
For the ARC, HellaSwag, MMLU, Winogrande, and GSM8K benchmarks, a 5-shot approach was applied, whereas a zero-shot setup was used for evaluating TruthfulQA. For evaluation metrics, we use acc\_norm for ARC and HellaSwag; acc for Winogrande, MMLU, and TruthfulQA; and strict exact\_match for GSM8K, aligning with the Open LLM Leaderboard.

The experimental results in Table~\ref{table:general_abilities} demonstrate that through key parameter mining and type-tailored tuning strategies, \textbf{\mname can significantly enhance adherence and robustness with minimal impact on the general capabilities of LLMs.}
In some cases, \mname even led to slight performance improvements. 
These findings provide fresh insights and inspiration for the broader applicability of \mname in diverse scenarios.

\begin{table}[t]
\centering
\small
\begin{tabular}{lc}
\toprule
\textbf{Models} &  Match Rate \\
\midrule
 LLaMA2-7B-Chat  & 1.36\%   \\
 Qwen1.5-14B-Chat & 1.89\%   \\
\bottomrule
\end{tabular}
\caption{\label{tab:memorization}The match rate between LLMs' parameterized answers and conflicting answers after training.}
\end{table}

\subsection{Sensitivity Analysis for Hyperparameters}
\label{appendix:sensitivity}
Our proposed \mname framework primarily includes two key hyperparameters: the smoothing hyperparameter \( \alpha \), used in Equation (\ref{eq:ema_uncertainty}) to calculate importance scores during backpropagation, and the reweighting hyperparameter \( \delta_1 \), applied in Equations (\ref{eq: loss-c}), (\ref{eq: loss-ax}), and (\ref{eq: loss-rx}) to balance the original tasks with the newly introduced tasks. We aim to explore how different settings of these hyperparameters affect the performance of \mname on the SQuAD2.0-Eval and RGB datasets, with tests conducted using the LLaMA2-7B-Chat backbone model.
As shown in Table~\ref{table:senstivity_alpha}, we observe that the optimal value for \( \alpha \) is 0.85. If the \( \alpha \) is set too low, both adherence and robustness experience a decline, indicating that the identification of important parameters is affected by random sampling and the complex dynamics of the training process.
Additionally, the results in Table~\ref{table:senstivity_delta} indicate that \(\delta_1\) values within a normal range have minimal impact on performance. Conversely, overly small \(\delta_1\) values can diminish the effectiveness of tasks aimed at enhancing model adherence and robustness. On the other hand, excessively large \(\delta_1\) values may undermine the positive effects of our proposed document extraction task.
Overall, the model maintains relatively stable performance under most conditions, indicating a low sensitivity to changes in hyperparameters.

\subsection{Detection of Unnecessary Memorization}
\label{appendix:memorize}

Introducing question-answer pairs and conflicting knowledge contexts (especially fictional knowledge and answers) into the training data presents a potential risk of the model developing unnecessary memory retention. We employ prompts that do not provide supplementary knowledge to extract the parameterized knowledge of LLMs fine-tuned with \mname. The results in Table~\ref{tab:memorization} demonstrate that the model retains almost none of the conflicting knowledge. This further confirms that \textbf{\mname effectively guides and optimizes the behavior patterns of LLMs across various contexts, without embedding specific knowledge into the model.}

\begin{figure}[t]
  \centering
    \includegraphics[width=1\linewidth]{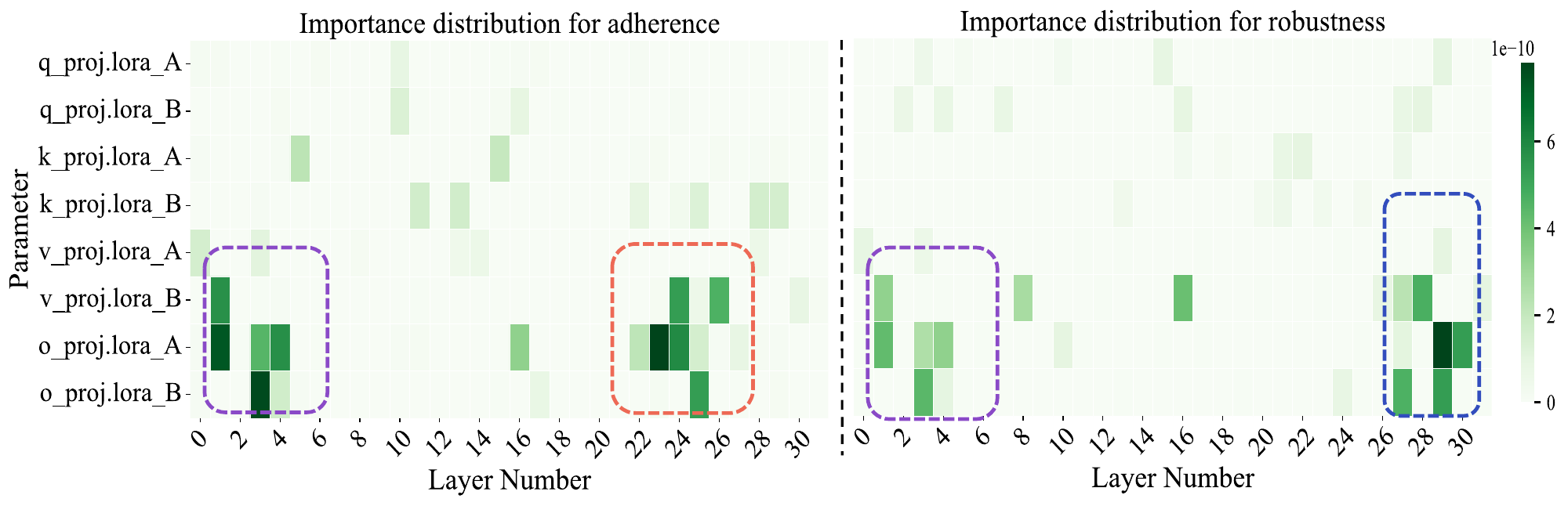}
  \caption{Visualization of parameter importance distributions $\mathcal{I}_a(\mathcal{E})$ and $\mathcal{I}_r(\mathcal{E})$ for adherence (\textbf{Left}) and robustness (\textbf{Right}) in Qwen1.5-14B-Chat.
  }
  \label{fig:heatmap-qwen1.5}
\end{figure}

\subsection{Additional Visualization Analysis}
\label{appendix:additional_vis}
In Figure~\ref{fig:heatmap-qwen1.5}, we visualize the distribution of parameter unit importance for adherence and robustness in Qwen1.5-14B-Chat.
Our observations align well with those detailed in Section~\ref{section:visualization_llama2}. Furthermore, in the Qwen1.5-14B-Chat model, we observe an increase in both entangled parameter units and robustness-specific units. This finding aligns with existing research~\cite{jin2024tug,xieadaptive}, which indicates that as language models grow in size, their capability to leverage internal memories for problem-solving also improves. Such enhanced reasoning abilities further augment LLMs' capacity to perceive external contexts effectively.

\subsection{More Diverse Evaluations}
\label{appendix:diverse_evaluations}
To further validate the generalizability of our approach to more complex real-world tasks, we perform additional experiments on the TempReason~\cite{tan2023towards} and MuSiQue~\cite{trivedi2021musique} dataset based on LLaMA2-7B-Chat.
\begin{itemize}[leftmargin=*,noitemsep,topsep=3pt]
\item \textbf{TempReason}~\cite{tan2023towards} contains three levels of time-related questions.It spans a time range from 634 to 2023 and includes a total of 52.8K question-answer pairs. We use this dataset to evaluate whether LLMs can select the most recent knowledge when confronted with conflicting answers from different timestamps. This is crucial for evaluating their performance in scenarios where knowledge bases are continuously updated.

\item \textbf{MuSiQue}~\cite{trivedi2021musique} is a challenging multi-hop question answering dataset designed to test the ability of LLMs to reason across all supporting facts in a given context to arrive at correct answers for complex questions. It contains 24,814 question-answer pairs, with questions involving 2 to 4 hops of reasoning.

\end{itemize}

\label{appendix:real-world-exp}
\begin{table}
\centering
\scalebox{0.9}{
\begin{tabular}{lcc}
\toprule
Method & TempReason & MuSiQue \\
\midrule
\rule{0pt}{6pt} Base & 20.90 & 43.71 \\
\rule{0pt}{8pt} KAFT & 28.70 & 58.72 \\
\rule{0pt}{8pt} \mname & \textbf{31.40} & \textbf{61.35} \\
\bottomrule
\end{tabular}}
\caption{\label{table:real-world-exp}Performance comparisons of \mname with other RAG methods (all utilizing the same LLaMA2-7B-Chat backbone) on complex real-world tasks. All results are reported in accuracy (\%).}
\end{table}

\begin{table}[t]
\centering
\begin{tabular}{lcccc}
\toprule
\multirow{2}{*}{Models} & \multicolumn{2}{c}{SQuAD} & \multicolumn{2}{c}{KNOT} \\
\cmidrule(lr){2-5}
                     & $R_{Ad}$          & $R_{Ro}$         & $R_{Ad}$         & $R_{Ro}$         \\
\midrule
Claude-3.5      &      68.29     &      29.76       &      63.85	      &      28.49      \\
GPT-4     &      69.16       &       28.21      &      64.24      &      27.56      \\
\mname   &        \textbf{80.47}     &       \textbf{47.58}      &      \textbf{73.27}      &      \textbf{43.34}      \\
\bottomrule
\end{tabular}
\caption{\label{table:closed_source}Performance comparisons of \mname (using Qwen1.5-14B-Chat as the backbone model) with closed-source LLMs Claude-3.5 Sonnet and GPT-4.}
\end{table}

The results in Table \ref{table:real-world-exp} indicate that our method \textbf{effectively supports real-world applications in dynamic scenarios and demonstrates strong generalization capability for more complex tasks. }

\begin{table*}[t]
\centering
\scalebox{1.0}{
\begin{tabular}{lccccc}
\hline
\multirow{2}{*}{Method} & \multicolumn{2}{c}{Probing phase} & \multicolumn{2}{c}{Training phase} & \multirow{2}{*}{Number of tokens}\\ 
& Time   & Memory usage & Time   & Memory usage &  \\ 
\hline
KAFT   & /      & /   & 1.49h  & 68.3GB       & 32.7M            \\ 
\mname & \textbf{154s}  & 68.3GB       & \textbf{1.03h}  & 68.3GB       & 33.5M  \\ 
\hline
\end{tabular}
}
\caption{\label{table:computation cost}Computational cost of \mname and the baseline KAFT on LLaMA2-7B-Chat.}
\end{table*}

\subsection{Comparison with Closed-Source LLMs}
\label{appendix:closed-source_LLMs}
To further validate the effectiveness of our approach, we conduct experiments on the SQuAD2.0-Eval and KNOT datasets, using the closed-source LLMs Claude-3.5 Sonnet\footnote{https://www.anthropic.com/news/claude-3-5-sonnet} and GPT-4~\cite{achiam2023gpt} for comparison.
The experimental results are shown in Table~\ref{table:closed_source}. Although these closed-source LLMs demonstrate strong performance, our proposed \mname (based on Qwen 1.5-14B-Chat) still exhibits superior knowledge control and selection capabilities when faced with complex retrieval contexts.

\subsection{Training Overhead and Computational Efficiency}
\label{appendix:computation_cost}
From the perspective of network propagation, the proposed \mname method mainly consists of two phases: key parameter mining and type-tailored tuning. We provide additional details regarding the workload involved in constructing the training dataset, as well as the training time and memory usage during the probing (key parameter mining) and training (type-tailored tuning) phases, and the number of training tokens. 

From results in Table \ref{table:computation cost}, we can observe that, under otherwise identical conditions, \mname, despite introducing an additional probing stage compared to baseline KAFT, actually results in \textbf{shorter training times and faster network convergence}. This suggests that our proposed strategy of "decoupling parameters first, then performing tailored tuning" effectively prevents conflicting supervisory signals from interfering with each other, thereby improving training efficiency and accelerating the process. Additionally, the total number of training tokens in our approach is comparable to that of KAFT, with no additional computational overhead. Furthermore, during the inference phase, the trained model performs with no difference in inference speed compared to the original base model.

\begin{table}[t]
\centering
\begin{tabular}{
  >{\centering\arraybackslash}m{2cm}
  >{\raggedright\arraybackslash}m{4.25cm}
}
\hline
{\textbf{Question}} & \textbf{Which animal is the fastest on land?} \\ \hline
\multirow{3}{*}{\textbf{Documents}}                 & [Doc1] The cheetah is considered the fastest land animal, reaching speeds of up to 70 miles per hour over short distances. \\ 
                                   & [Doc2--noise] The pronghorn antelope, native to North America, can reach speeds of 55 miles per hour, making it one of the fastest land animals. \\ \hline
\textbf{Parenting Answer}          & Based on extra knowledge, the cheetah is the fastest land animal. \textbf{[CORRECT]} \\ \hline
\textbf{KAFT Answer}               & Based on extra knowledge, the pronghorn antelope is the fastest land animal. \textbf{[WRONG]} \\ \hline
\end{tabular}
\caption{\label{table:case study}Case study comparing the baseline KAFT and \mname.}
\end{table}

\subsection{Case Study}
\label{appendix:case_study}
In this section, we provide a case study to visually demonstrate the effectiveness of \mname.
As shown in Table \ref{table:case study}, the two external knowledge documents we provided are semantically highly similar to the question. However, the noise document does not contain a direct answer, as it only describes the ranking of antelope speed using the term "one of", without providing a precise answer.

The baseline KAFT method treats adherence and robustness as analogous instruction-following positive examples. This approach introduces contradictory signals during instruction-following training, leading to increased learning variance and impeding the effective acquisition of both capabilities. As a result, KAFT fails to comprehensively analyze the relevance between the context and the question at a fine-grained level, making it more susceptible to noise interference, ultimately leading to incorrect answers.

In contrast, Parenting effectively addresses the balance between adherence and robustness by decoupling and identifying parameter subspaces associated with each capability. By designing specialized tuning strategies for these distinct subspaces, Parenting ensures that both adherence and robustness are fully optimized. As a result, it can effectively leverage evidence to generate the correct answer.

\end{document}